\documentclass[letterpaper, 10 pt, journal, twoside]{IEEEtran}
\usepackage{url}
\usepackage{svg}
\usepackage{hyperref}
\usepackage[backend=biber,
            hyperref=true,
            url=false,
            isbn=false,
            doi=false,
            backref=false,
            style=ieee,
            natbib=true,
            mincitenames=1,
            maxcitenames=1,
            citestyle=numeric-comp,
            sorting=nyt,
            block=none]{biblatex}

\bibliography{bibliography}
\usepackage{tabularx}
\usepackage{amsmath}
\usepackage{amssymb}
\usepackage{amsfonts}
\usepackage{mathtools}
\usepackage{xcolor} 
\usepackage{graphicx}
\usepackage{caption}
\usepackage{subcaption}
\usepackage{makecell}
\usepackage[a-1b]{pdfx}
\usepackage{color, soul} 

\newcommand{\fig}[1]{Fig.~\ref{#1}}  
\newcommand{\tbl}[1]{Table~\ref{#1}}

\begin{document}
\markboth{IEEE Robotics and Automation Letters. Preprint Version. July, 2022}
{Qi \MakeLowercase{\textit{et al.}}: Dough Manipulation} 
\title{
Learning Closed-loop Dough Manipulation Using a Differentiable Reset Module}
\author{Carl Qi$^{1}$, Xingyu Lin$^{1}$, David Held$^{1}$
\thanks{Manuscript received: February 25, 2022; Revised: May 21, 2022; Accepted: June 16, 2022.}%
\thanks{This paper was recommended for publication by Editor Markus Vincze upon evaluation of the Associate Editor and Reviewers’ comments.}%
\thanks{This material is based upon work supported by the National Science Foundation under Grant No. IIS-2046491, IIS-1849154 and LG Electronics. We thank Zhiao Huang for helping us with the simulator and thank Tim Angert and Sarthak Shetty for helping us with real world experiments.}
\thanks{Carl Qi, Xingyu Lin, and David Held are with the Robotics Institute, Carnegie Mellon University, Pittsburgh, PA, USA. \tt\footnotesize hanwenq@andrew.cmu.edu, xlin3@andrew.cmu.edu, dheld@andrew.cmu.edu.}%
\thanks{Digital Object Identifier (DOI): see top of this page.}%
}
\maketitle

\begin{abstract}
Deformable object manipulation has many applications such as cooking and laundry folding in our daily lives. Manipulating elastoplastic objects such as dough is particularly challenging because dough lacks a compact state representation and requires contact-rich interactions. We consider the task of flattening a piece of dough into a specific shape from RGB-D images. While the task is seemingly intuitive for humans, there exist local optima for common approaches such as naive trajectory optimization. We propose a novel trajectory optimizer that optimizes through a differentiable ``reset" module, transforming a single-stage, fixed-initialization trajectory into a multistage, multi-initialization trajectory where all stages are optimized jointly. We then train a closed-loop policy on the demonstrations generated by our trajectory optimizer. Our policy receives partial point clouds as input, allowing ease of transfer from simulation to the real world. We show that our policy can perform real-world dough manipulation, flattening a ball of dough into a target shape. Supplementary videos can be found on our project website: \footnotesize \url{https://sites.google.com/view/dough-manipulation}.

\end{abstract}

\section{INTRODUCTION}


\IEEEPARstart{D}{eformable} object manipulation allows autonomous agents to expand their applicability in our daily lives. Many household tasks such as folding laundry, cleaning rooms, and cooking food require a substantial amount of interaction with deformable objects, and recent works~\cite{2020vsf, Wu-RSS-20, lin2021learning, ha2022flingbot, kim2022planning, foodmanip} have shown great progress towards building a household robot. However, there are many challenges within deformable object manipulation that are yet to be solved. For one, deformable objects lack a compact state representation for manipulation. The state of rigid objects can often be captured by their 6D poses~\cite{nagabandi2020deep, chen2021system, underactuated}. On the other hand, deformable objects have high degrees of freedom, making them hard to model. Many prior works rely on hand-designed representations for deformable object manipulation~\cites{kim2022planning,Matl2021DeformableEO,pizza}, which results in task-specific representations that lack generalizability. Recent works design more general data-driven methods to directly learn a policy from RGB-D images~\cites{Wu-RSS-20, Matas2018SimtoRealRL} or a dynamics model~\cite{yan2020learning, 2020vsf, li2018learning, lin2021learning}. These works often directly interact with the deformable objects with the robot gripper and use pick-and-place actions as the primitive. However, elastoplastic object manipulation is particularly challenging, as they require rich contact for interacting with the object~\cite{ruggiero2018nonprehensile}, often with an external tool. Simple action primitives are not sufficient in these cases. In this work, we tackle the challenges of elastoplastic object manipulation without the aid of feature engineering or action primitives.

\begin{figure}[t]
    \centering
    \includegraphics[width=\linewidth]{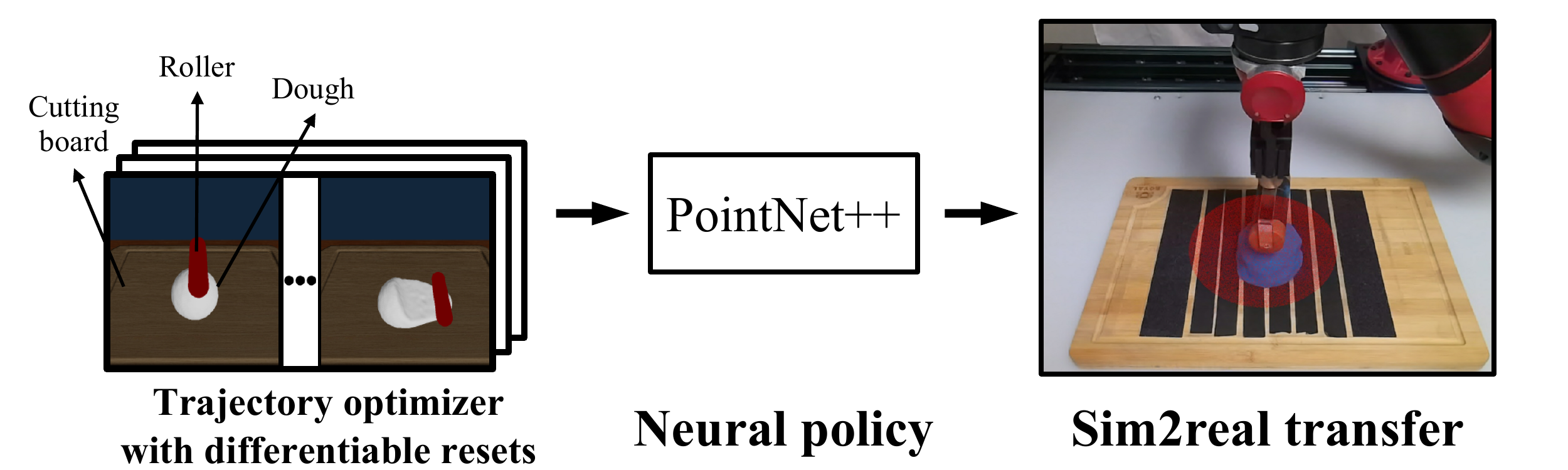}
    \caption{\small We use a novel trajectory optimizer to generate demonstration data in simulation. We then use this demonstration data to train a closed-loop point cloud based policy using PointNet++ \cite{qi2017pointnetplusplus} as an encoder. Our policy can transfer to the real world without any fine-tuning.}
    \label{fig-intro}
    \vspace{-7mm}
\end{figure}


Manipulation of elastoplastic objects like dough and clay has wide application for cooking and art-making, and there exists a rich list of literature on food manipulation~\cite{foodmanip,bollini2013,tokumoto2002deformation, pizza, lin2022diffskill}. Similar to other deformable objects, prior works have proposed action primitives for dough flattening~\cite{kim2022planning,tokumoto2002deformation,Matl2021DeformableEO}. They parameterize the action space by a few parameters such as the direction and length of a rolling trajectory, which limits the flexibility and efficiency of the obtained trajectory. In contrast, we learn closed-loop control policy by imitating a trajectory optimizer, allowing the robot to learn more complex motions without limiting it to one type of manipulation.
\begin{figure*}[t]
    \centering
    \includegraphics[width=0.7\textwidth]{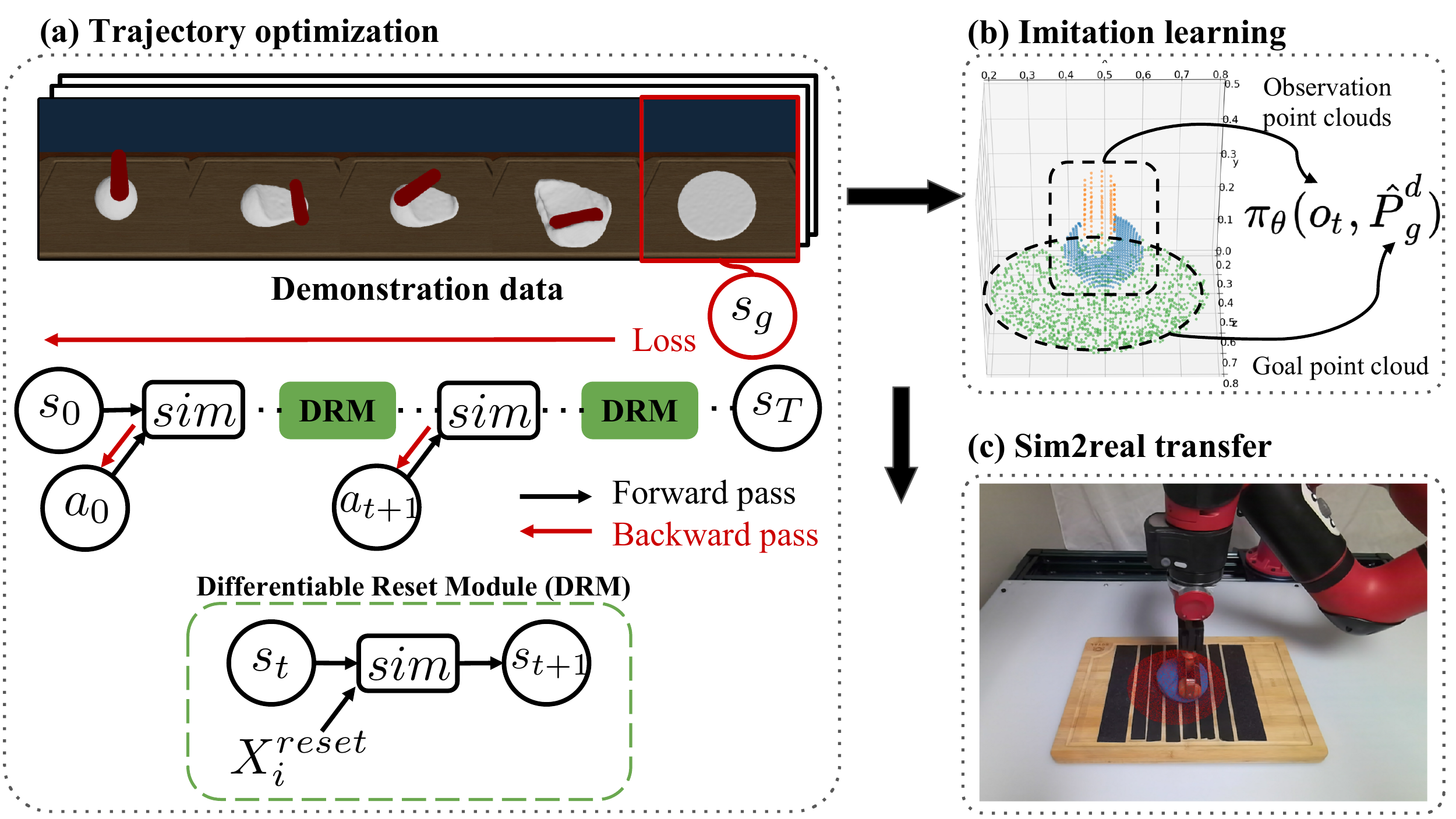}
    \caption{\small An overview of our method. (a) Our trajectory optimizer implements a differentiable module that resets the tool to avoid local optima and allows gradient information to propagate through the entire trajectory. (b) We perform imitation learning on the demonstration data generated by the trajectory optimizer. Our policy takes segmented point cloud observations as input.  (c) We use the learned policy to control a sawyer robot to perform closed-loop rolling in the real world.
    }
    \label{fig-overview}
    \vspace{-5mm}
\end{figure*}

Enabled by recent progress on differentiable simulators~\cite{hu2019difftaichi, heiden2021disect,huang2021plasticinelab}, gradient-based trajectory optimization (GBTO) allows us to acquire many manipulation skills. However, many tasks require multistage manipulation and running GBTO from a fixed state initialization can result in locally optimal solutions.
Solving these multi-stage tasks also require long-horizon reasoning. Prior work has proposed learning abstraction of the skills for planning over multi-stages~\cite{lin2022diffskill}. In this work, we propose a simpler method to get around the local optima in GBTO by adding a differentiable reset module (DRM). Our key observation is that, we can often identify a few reset poses for the robot to be in between different stages of manipulation to help GBTO get around local optima. Consider the task of flattening a piece of dough into a circle. It is hard to perform the task with one simple rolling motion, and if one tries to roll multiple times, overly flattening the dough in early rolls will cause later rolls to break the dough. As such, this seemingly easy task actually requires joint optimization of a multi-stage motion. To tackle these issues, DRM samples a few reset poses for the roller to be after each stage and differentiate through the reset motion to joinly optimize the multi-stage motion. Finding these reset poses is often easy and can significantly improve the performance of GBTO. We give another such example in cooking: when seasoning a pizza, the optimal motion requires tilting the shaker multiple times, each at a different location, instead of dumping all seasonings in one spot.
 


Finally, we use our trajectory optimizer in combination with a differentiable simulator to efficiently perform gradient-based optimization and generate high-quality demonstrations, as similarly done in~\cite{lin2022diffskill}. We can then train a policy via imitation learning with partial point clouds as input. Our point cloud based policy is robust to occlusion in the scene and can effectively flatten a piece of dough in a smooth and closed-loop fashion.
To summarize, in this paper:
\begin{itemize}
    \item We propose a DRM in GBTO that optimizes a multistage trajectory end-to-end and avoids local optima.
    \item To the best of our knowledge, we introduce the first system for closed-loop dough manipulation in the real world. Our system operates directly from partial point cloud and does not require any feature engineering or action primitives.
\end{itemize}

\section{RELATED WORK}


\subsection{Model-based approaches for dough manipulation}
Many works use a model-based approach for dough manipulation. Tokumoto et al.~\cite{tokumoto2002deformation} proposes an analytical dynamics model for flattening a piece of dough under a specific action primitive. The rolling primitive is a straight line motion of the roller with a fixed height and angle, where the height is predefined at test time, so the task becomes a simple 1D optimization over the rolling direction. Moreover, the analytical dynamics model is tailored to the rolling primitive, which makes it unusable for any other action parameterization or tasks. Recent works \cite{kim2022planning,Matl2021DeformableEO,task-movement13,pizza, li2018learning} also use action primitives for their tasks but learn a dynamics model in a data-driven fashion to improve generalization. However, many rely heavily on feature engineering for the state of the dough: Kim et al.~\cite{kim2022planning} uses the set of distances between the boundary points and the center of the dough, Matl et al.~\cite{Matl2021DeformableEO} uses a bounding box for the dough, and Calinon et al. and Figueroa et al.~\cite{task-movement13,pizza} fit an ellipse on the 2D view of the dough. These manually defined state spaces have limited representation power and cannot easily transfer to other tasks. Among them, DPI-Net \cite{li2018learning} learns a particle dynamics model that functions over point clouds and does not require a manually specified state representation. However, DPI-Net requires full point cloud observation acquired by moving the robot arm out of the scene after each action, which significantly slows down their execution. Furthermore, DPI-Net uses a derivative free planner that also suffers from the local optima in our task. To compare with DPI-Net, we evaluate a CEM planner with ground-truth dynamics and reward in simulation. We also compare with a ``heuristic'' baseline that uses a similar action primitive described in ~\cite{tokumoto2002deformation,kim2022planning,Matl2021DeformableEO,task-movement13} in the real world to show the advantage of a more flexible action space. 
Last, Figueroa et al.~\cite{pizza} does not specify a fixed rolling primitive; instead, they automatically discover a set of dynamical systems (DS) from human demonstration, which they later use to perform rolling. Despite the advantages of DS in contact-rich interactions~\cite{smoothcontact,dsjournal}, a good dynamical system for rolling is difficult to learn, and as a result, they require a significant number of rolls (between 12 and 71) to achieve the goal.
In contrast to these works, we do not simplify the control by specifying a rolling primitive, nor do we manually define a compact state representation. These choices make the dough flattening problem significantly harder but meanwhile make our approach applicable to a general type of manipulation.
\vspace{-3mm}
\subsection{Model-free deformable object manipulation}
As opposed to learning a dynamics model, another way is to learn a policy that directly outputs the actions. Wu et al.~\cite{Wu-RSS-20} and Matas et al.~\cite{Matas2018SimtoRealRL} learn a policy via Reinforcement Learning (RL). However, in our experiments with a Soft Actor Critic (SAC)~\cite{haarnoja2017soft} agent, we discover that RL with high dimensional input results in unmeaningful behavior. We include it as a baseline.
Seita et al.~\cite{seita_fabrics_2020} and Lin et al.~\cite{lin2022diffskill} perform Imitation Learning (IL) with either an algorithmic supervisor or a trajectory optimizer that has access to privileged state information. Comparing to Seita et al.~\cite{seita_fabrics_2020}, which uses a algorithmic supervisor to a specific problem, our approach is more general. Similar to Lin et al.~\cite{lin2022diffskill}, we train our policy via Behavioral Cloning (BC) from the demonstrations generated by a trajectory optimizer, but our point cloud representation as policy input allows us to perform sim2real transfer with no fine-tuning.

\subsection{Differentiable simulators for trajectory optimization}

As deformable object manipulation becomes increasingly popular, the need for training data results in many high-quality simulators \cite{hu2019difftaichi,corl2020softgym, heiden2021disect,huang2021plasticinelab}. We choose PlasticineLab~\cite{huang2021plasticinelab}, which uses Material Point Method~\cite{hu2018moving} to model elastoplastic material. Built on top of the DiffTaichi system~\cite{hu2019difftaichi}, it supports differentiable physics that allows us to perform gradient-based trajectory optimization.

There exists many prior works on non-convex trajectory optimization. Differential Flatness~\cite{Murray95differentialflatness} exploits the fact that sometimes a dynamical system's state and control variables can be represented purely as a function of the system's output and its derivatives, and thus the state and control variables can be solved efficiently. Iterative LQR~\cite{ilqr} performs a linear and quadratic approximation to the system's dynamics and cost function and iteratively refines the trajectory. However, with a large state space, finding a differentially flat system or approximating a nonlinear dynamics are both difficult.
On the other hand, the differentiable simulator has the ability to compute the gradient information of the next state with respect to the previous state and action. Thus, using the backpropagation technique in neural network training, one can optimize a trajectory within the differentiable simulator with gradient descent~\cite{huang2021plasticinelab}.
However, as pointed out in Li et al.~\cite{li2022contact}, this type of trajectory optimization suffers from local optima in slightly more complicated tasks, and they mitigate this issue via contact point discovery. Our approach is orthogonal to~\cite{li2022contact} and can be used in combination with contact point discovery to find better reset poses. Most relevant to our work is DiffSkill~\cite{lin2022diffskill}, which divides a long-horizon problem into simpler subproblems and leverages learned action primitives for long-horizon planning. However, DiffSkill uses more complex planning method and has only been shown to work in simulation with RGB-D images, which makes it hard to transfer to the real world. In contrast, our method uses simple reset poses to get around local optima and inputs partial point cloud for the policy, enabling easier sim2real transfer.
\begin{figure}[t]
    \centering
     \begin{subfigure}[b]{\linewidth}
         \centering
         \includegraphics[width=0.8\linewidth]{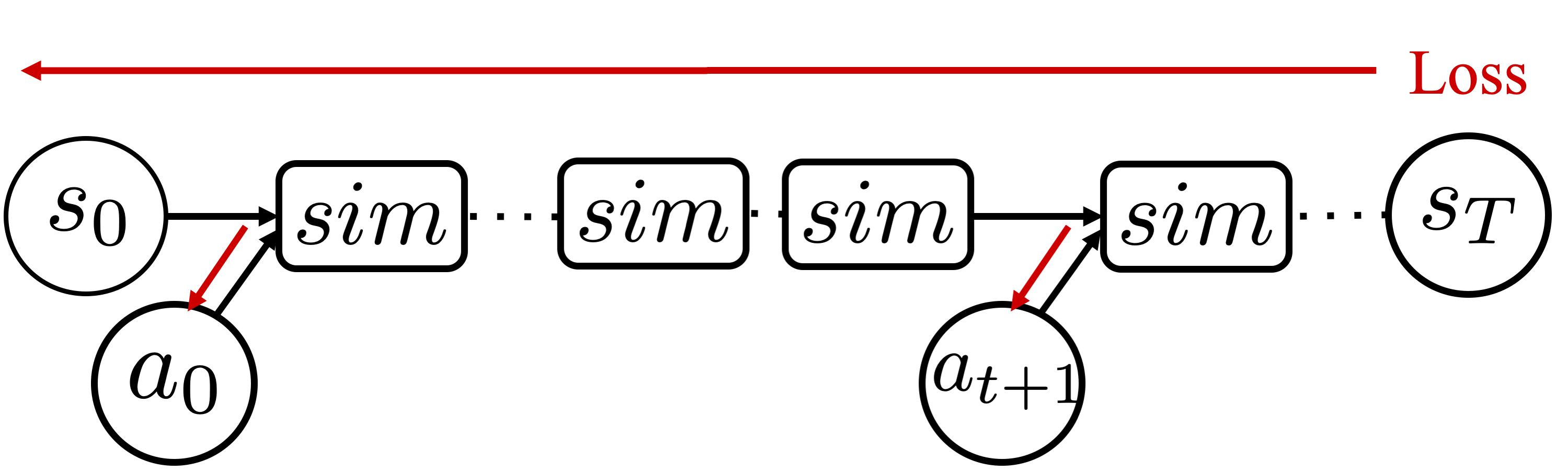}
      \centering
      \caption{Naively optimize the entire trajectory.}
      \label{fig-traj-naive}
     \end{subfigure}
         
     \begin{subfigure}[b]{\linewidth}
     \centering
     \includegraphics[width=0.8\linewidth]{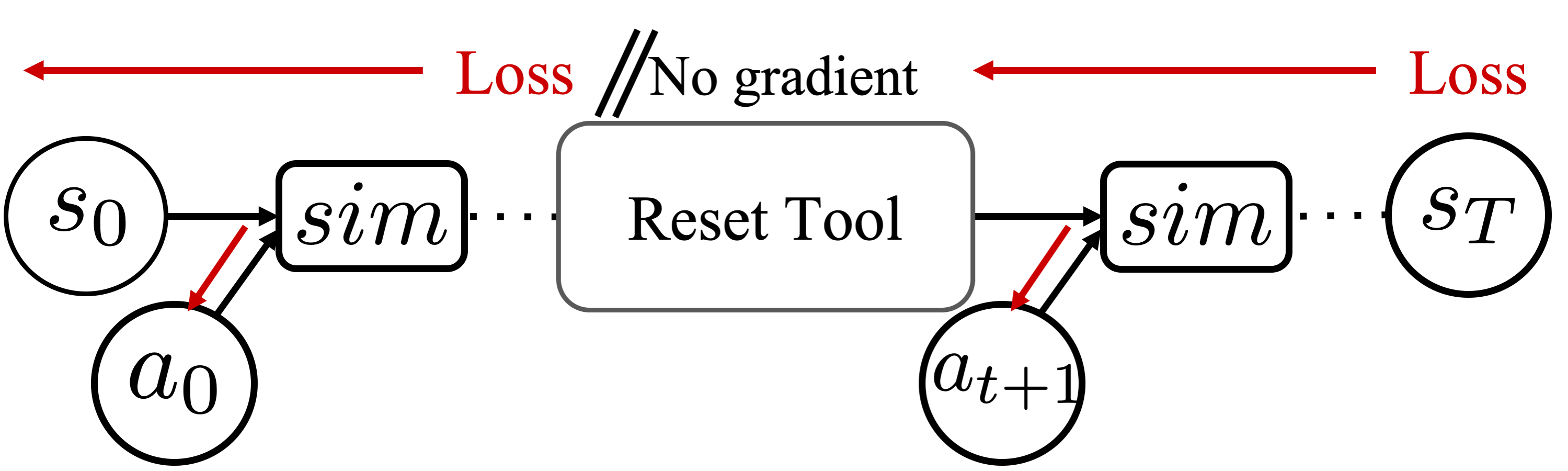}
         \caption{Optimize multiple trajectories separately with tool resets in b/w.}
         \label{fig-traj-sep}
     \end{subfigure}
          \begin{subfigure}[b]{\linewidth}
     \centering
     \includegraphics[width=0.8\linewidth]{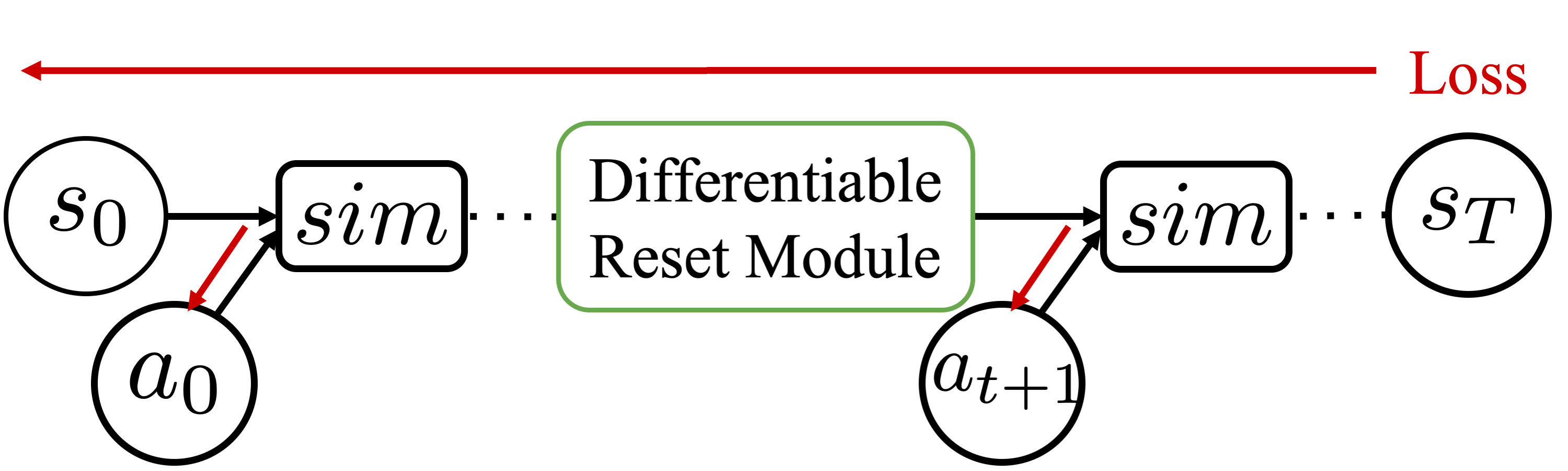}
         \caption{\small Optimize the entire trajectory with differentiable reset modules.}
         \label{fig-traj-ours}
     \end{subfigure}

    \caption{\small Comparison of different trajectory optimization methods discussed in~\ref{method-b}. (\ref{fig-traj-naive}) suffers from local optima; Gradient in (\ref{fig-traj-sep}) cannot propagate from one trajectory to another; (\ref{fig-traj-ours}) allow us to optimize a multistage trajectory.
    }
    \label{fig-traj}
\vspace{-5mm}
\end{figure}

\section{Problem formulation}
We formulate our dough manipulation task into a Partially Observable Markov Decision Process (POMDP) with state space $\mathcal{S}$, action space $\mathcal{A}$, observation space $\Omega$, deterministic transition dynamics $\mathcal{T}: \mathcal{S} \times \mathcal{A} \rightarrow \mathcal{S}$, and mapping from states to observations $O: \mathcal{S} \rightarrow \Omega$. In our simulation, we have access to the ground truth state information $s$, which includes the ground-truth dough particles $P^d \subseteq \mathcal{R}^3$ as well as the pose of the tool $X^t \in SE(3)$. The robot observation $o_t := \hat P^d_t \oplus P^t_t$ is the partial point cloud of the dough captured by a depth camera ($\hat P^d_t)$ concatenated with the ground-truth point cloud of the tool ($P^t_t$), which we obtain from our prior knowledge of the tool shape and the end-effector position. The robot also observes the target dough point cloud $\hat P^d_g$ specified by a human. We assume that our transition function $\mathcal{T}$ is differentiable, which can be achieved using either a learned dynamics model~\cite{li2018learning} or a differentiable simulator, such as PlasticineLab~\cite{huang2021plasticinelab}.
\section{METHOD}
Our pipeline consists of three main components: generating demonstrations via a novel trajectory optimizer, training a point-cloud based policy, and transferring the policy to the real world. For our trajectory optimizer, we leverage a reset module that avoids local optima and allows the gradient to back-propagate from one end of an episode to the other. We then use the demonstrations generated by the trajectory optimizer to train a policy that takes in partial point clouds as input. Last, we transfer the policy by mapping the action space from simulation to the real world. We discuss each of the components below in detail.


\subsection{Demonstrations from non-local trajectory optimization}
\label{method-b}
To generate expert demonstrations for rolling, we leverage a gradient-based trajectory optimization~(GBTO) with the differentiable simulator. We write our trajectory optimization in Direct Shooting form~\cite{underactuated} with all the constraints captured by the dynamics of the environment. Thus, given the initial state $s_0$, the objective is:
\vspace{-5mm}
\begin{flalign}
    \min_{\substack{a_1, ..., a_T \\\text{where}~s_{t+1} = \mathcal{T}(s_t, a_t)}} &L_{traj} = \sum_{t=0}^T L^{roll}_t + \lambda L^{contact}_t \label{eq:loss}
\end{flalign}
 where $L^{roll}$ measures the Earth Mover's Distance ($D_{EMD}$)~\cite{emd} between the current and target dough shapes. $L^{contact}$ measures the distance from the tool to the closest particle to encourage the tool to approach the dough and $\lambda$ weighs the different loss terms. This is the same objective used for dough manipulation in prior work DiffSkill~\cite{lin2022diffskill}.

With differentiable dynamics, we can back-propagate the gradient of the loss w.r.t the actions as follows:
\begin{align}
    \frac{\partial L_{traj}}{\partial a_t} = \sum_{t'=t+1}^T \frac{\partial \left(L^{roll}_t + \lambda L^{contact}_t\right)}{\partial s_{t'}} \frac{\partial s_{t'}}{\partial a_t}
    \label{eq:gradient_updates}
\end{align} 
where $L^{roll}_t$ and $L^{contact}_t$ are differentiable w.r.t. $s_t$. A straightforward approach used by many others~\cite{huang2021plasticinelab} is to use gradient descent to directly update the actions, as shown in \fig{fig-traj-naive}. However, gradient descent is a local optimization method, and the resulting action sequence will fall into a local optimum; specifically, the contact loss $L^{contact}_t$ will cause the tool to stay in close contact with the dough, which will cause the policy to perform only one rolling action (even if multiple rolling actions are needed to achieve the goal).
Instead of using complex planning methods~\cite{lin2022diffskill}, we notice that a simpler approach is to divide an episode into multiple rolling trajectories with tool resets in between. Each trajectory has its own tool initialization and would be optimized separately w.r.t. $L_{traj}$. These reset poses break out of the local optima resulted from a fixed initialization of the tool. However, as mentioned earlier, flatting a piece of dough requires joint optimization of multiple rolls. As shown in \fig{fig-traj-sep}, simply dividing up an episode will break the gradient flow from one rolling trajectory to another, which will lead to a suboptimal solution. 


Ideally, we would like to reset the tool in between different rolls and jointly optimize these rolls. To do so, we propose a novel trajectory optimization method that implements a differentiable reset module (DRM), as illustrated in \fig{fig-traj-ours}. The DRM can replace any action within a trajectory with a tool reset. We then can pick a set of timesteps $\{t^{reset}_1, ..., t^{reset}_k\}$ for DRM to move tool to a set of reset poses $\{X^{reset}_1, ..., X^{reset}_k\}$. Effectively, the timesteps separated by the DRMs are the different stages of rolling, and the DRMs provide multiple initializations of the tool in a trajectory.
During trajectory optimization, we keep the rest of the gradients the same but do not back-propagate the gradient from the reset poses, i.e. we set $\frac{\partial L_{traj}}{\partial X^{reset}_{i}} = 0, \forall i \in \{1, ..., k\}$. For the dough
flattening task, we find that it is easy to get around the local optima by specifying a few reset poses inspired by a human rolling dough. In general, our method is also useful to other tasks where it can be easy to specify a few intermediate reset poses. Using this approach, we transform a single-stage, fixed-initialization trajectory into a multistage, multi-initialization trajectory where all stages are optimized jointly. Our optimizer avoids local optima and produces synergistic behaviors between multiple rolls, as later shown in Section~\ref{sec-exp}. A detailed illustration of our trajectory optimizer is shown in \fig{fig-overview}.
\subsection{Point-cloud policy learning via imitation learning}
We deploy the above trajectory optimization method  to generate demonstration trajectories, which we will use to train a policy from partial point clouds, as described in this section.  Specifically, we use the trajectory optimization method described above to create a demonstration dataset  $\mathcal{D} = \{o^{(i)},  a^{(i)}, \hat P^{d^{(i)}}_g \}_{i}$, where $i$ indexes over trajectories.  We use this data to train a goal-conditioned policy with hindsight relabeling via imitation learning~\cite{lin2022diffskill}.
Our policy $ \hat a_t = \pi_{\theta}(o_t, \hat P^d_g)$
takes in the partial point cloud of the dough and ground-truth point cloud of the tool $o_t$, as well as the partial point cloud of the target dough shape $\hat P^d_g$, and outputs the action $\hat a_t$. Using a point cloud as input to our policy can help minimize the sim2real gap when transferring the policy to the real world. We train our policy using standard behavior cloning~(BC) with the following loss:
\vspace{-1mm}
\begin{align}
    L_{BC} = E_{(o, a, \hat P^d_g)\sim \mathcal{D}} \left[ \|a - \pi_{\theta}(o, \hat P^d_g)\|^2 \right]
\end{align}
Details on our policy architecture are discussed in Appendix~\ref{app-policy}.

\begin{figure}[t]
    \centering
    \vspace{3mm}
    \includegraphics[width=0.4\linewidth]{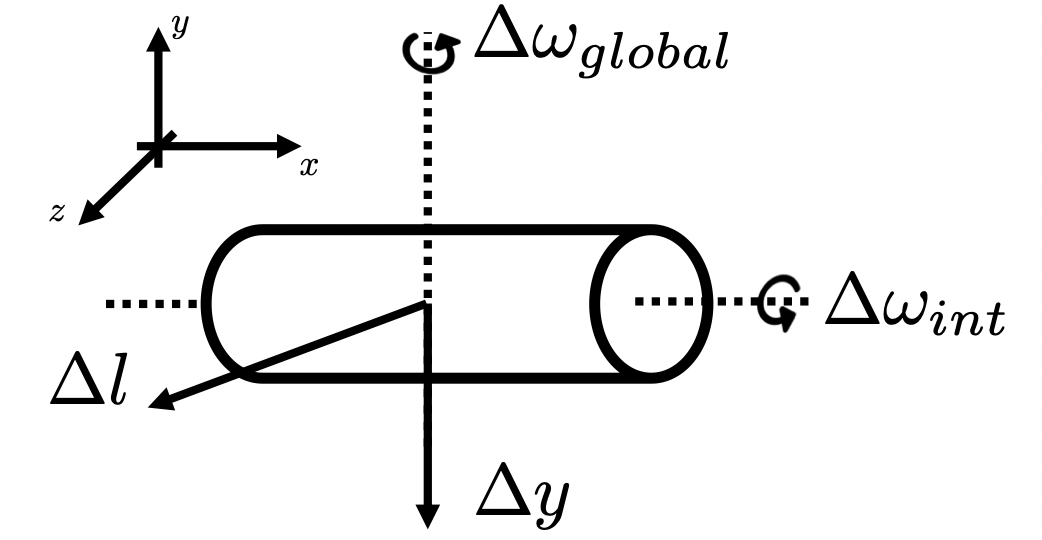}
    \begin{tabular}{ |c|c| }
    \hline
     Action & Description \\
     \hline
     $\Delta y$ & displacement of roller height\\
     \hline
     $\Delta l$ & displacement along rolling direction \\ 
     \hline
     $\Delta \omega_{int}$ & internal rotation of the roller \\
     \hline
     $\Delta \omega_{global}$ & rotation around the vertical axis \\
     \hline
    \end{tabular}
    \caption{\small Action space of the simulation environment.}
    \label{fig-actionspace}
\vspace{-5mm}
\end{figure}

\subsection{Sim2real transfer}

To transfer the policy to the real world, we parameterize the action space to abstract away the low-level controller. Specifically, given the policy output $\hat a_t$ in simulation, we use the environment's forward dynamics $\mathcal{T}$ to compute the target pose of the tool $\hat X_{t+1}$ and then apply a transformation on the coordinate system to obtain the real world target pose $\hat X^{real}_{t+1}$. Then we solve for the target robot joint angles with inverse kinematics and perform position control. We obtain the segmented dough point cloud by color thresholding and obtain the tool pose from the proprioceptive information of the robot. We then sample points uniformly on the surface of the tool to obtain the tool point cloud.  More details on our real world setup in detail can be found in Section~\ref{exp-realworld}.

\subsection{Implementation details}
\label{method-detail}
Our trajectory optimizer runs Adam Optimizer~\cite{adam} for $1000$ steps with a learning rate of $0.005$. We use a single reset module with a total of 100 rolling timesteps for our task because results in Section~\ref{exp-ablation} suggests this combination is the most effective. For other tasks, one can adjust the number of resets and time horizon accordingly. We use our trajectory optimizer to generate $150$ demonstration trajectories uniformly sampled over $125$ initial and target configurations with different sizes and locations of the dough. We then train our policy using behavior cloning and add Gaussian noise $\epsilon \sim \mathcal{N}(0, 0.01)$ to the partial point clouds during training to prevent overfitting. 
Our policy consists of a standard PointNet++ \cite{qi2017pointnetplusplus} encoder followed by a MLP. Details on the policy architecture can be found in Appendix~\ref{app-policy}.

\section{EXPERIMENTS}
\label{sec-exp}
Our experiments aim to demonstrate the effectiveness of our pipeline for dough manipulation. Specifically, we show that our trajectory optimizer outperforms the existing alternatives and produces high-quality demonstrations in simulation. We then evaluate our policy that uses partial point clouds as input, in both simulation and the real world. 

\begin{figure*}[t]
     \centering
     \begin{subfigure}[b]{0.65\textwidth}
         \centering
         \includegraphics[width=\textwidth]{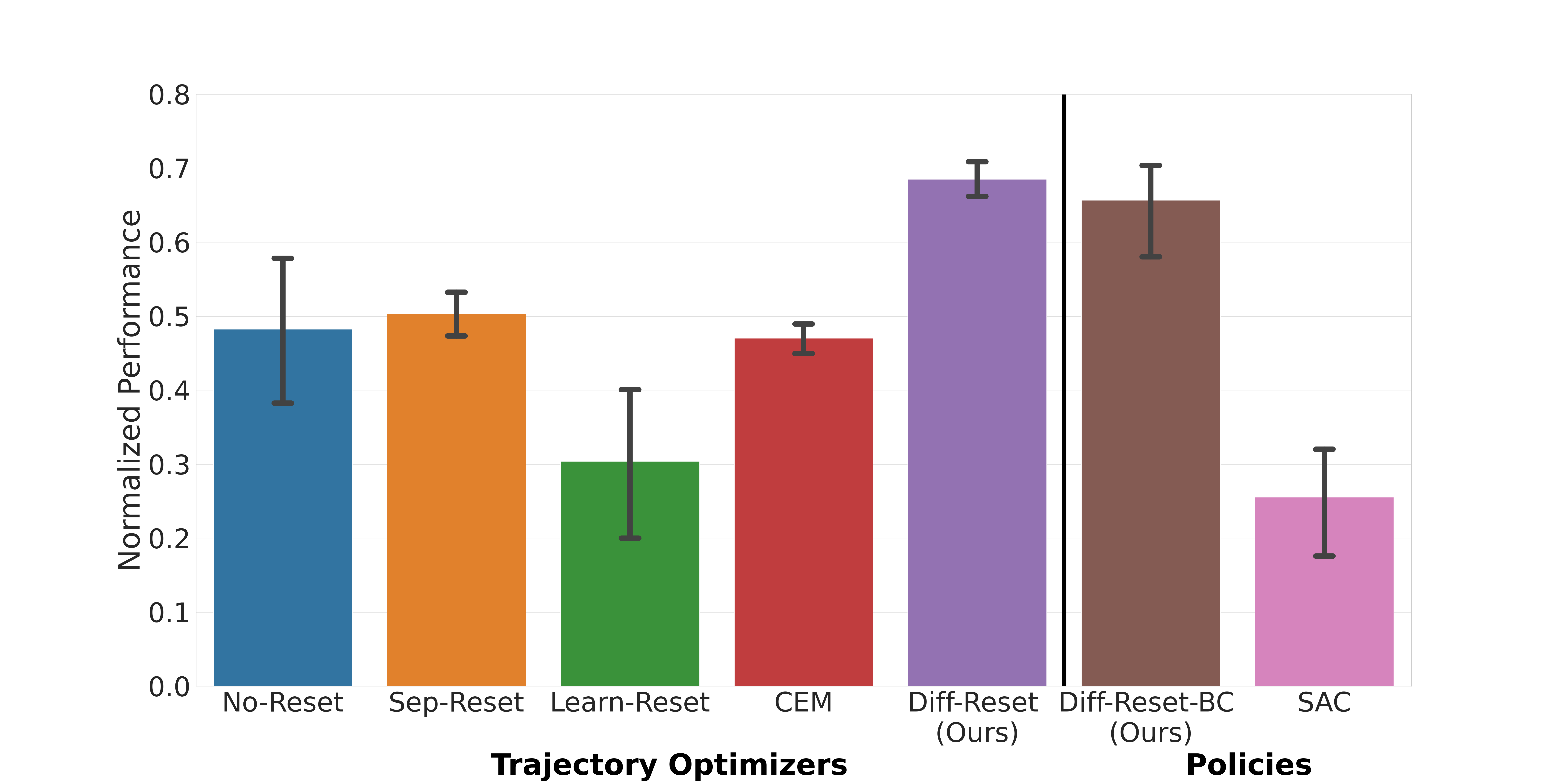}
         \caption{Normalized performance on 10 held out configurations. The performance of \\ the SAC agent is averaged over 4 random seeds.}
         \label{fig-main-result1}
     \end{subfigure}
     \hfill
     \begin{subfigure}[b]{0.33\textwidth}
         \centering
         \includegraphics[width=\textwidth]{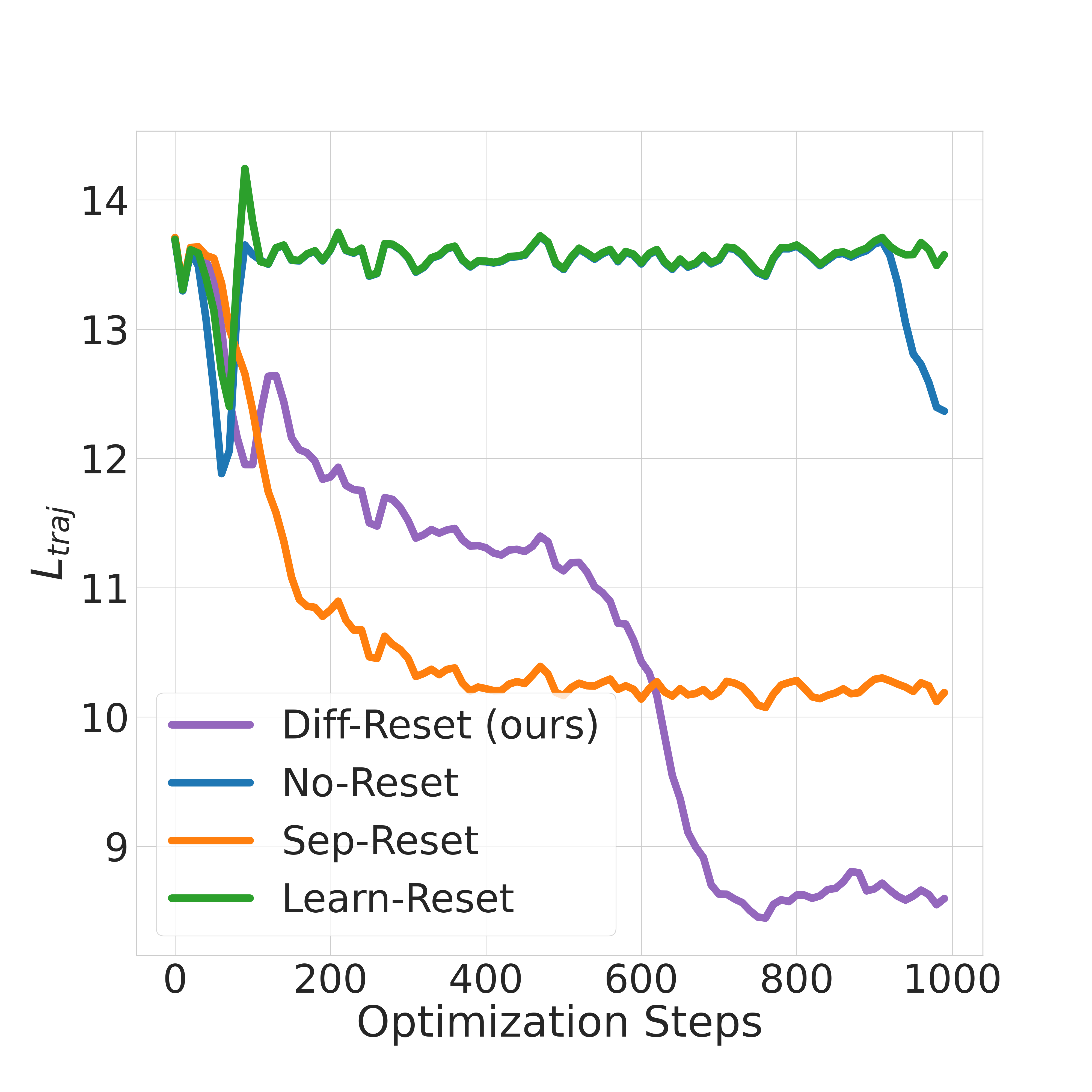}
         \caption{Loss $L_{traj}$~of different trajectory optimizers over optimization steps on an example trajectory.}
         \label{fig-main-result2}
     \end{subfigure}
     \hfill
     \caption{\small Results of simulation experiments. (\ref{fig-main-result1}) shows the performances of our method comparing to all baselines and ablations. (\ref{fig-main-result2}) shows the trajectory loss of different trajectory optimizers over time.}
     \label{fig-main-result}
     \vspace{-3mm}
\end{figure*}

\begin{figure}[t]
    \centering
    \includegraphics[width=\linewidth]{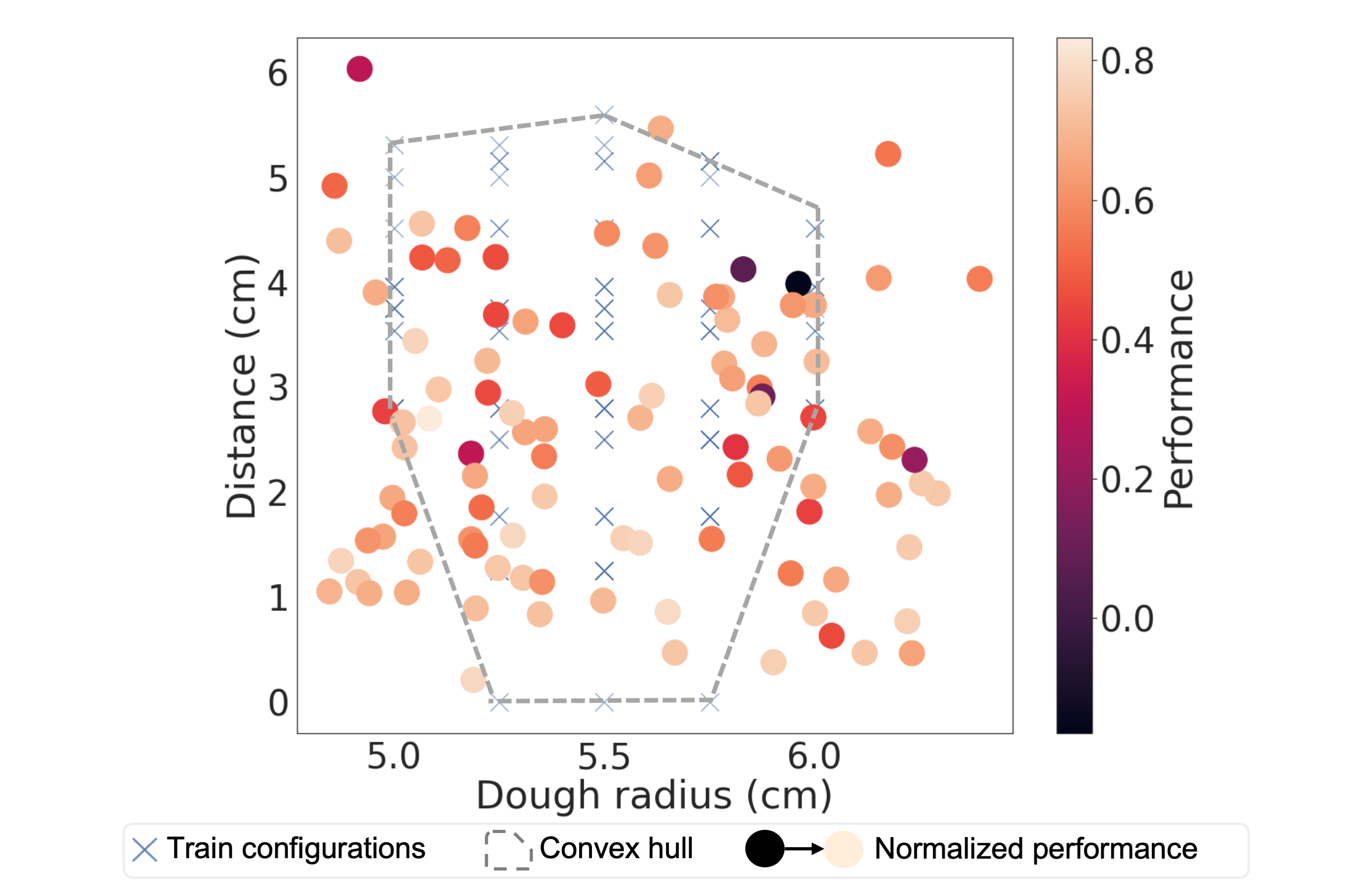}
    \caption{\small Diff-Reset-BC Policy performance over different configurations of dough radius and target distance. Train configurations (`X') and the convex hull (grey dotted lines) for training data are overlayed. Our policy can effectively extrapolate to unseen configurations.
    }
    \vspace{-5mm}
    \label{fig-heatmap}
\end{figure}

\subsection{Experiments setup} 
\textbf{Task.} We conduct our simulation experiments in PlasticineLab \cite{huang2021plasticinelab}, which provides a differentiable simulator that can simulate elastoplastic materials such as dough. Given a dough in a spherical shape, our task is to use a cylindrical roller to flatten the dough into a circular shape. We actuate the roller in simulation using a 4-dimensional continuous action space, as described in \fig{fig-actionspace}. The actions input to the simulator aim to specify the infinitesimal rigid transformation of the roller at every simulation step.
We vary the initial and target dough locations, initial roller location, as well as the volume of the dough. An example of a trajectory in our simulation environment is shown at the top of \fig{fig-overview}. 

\textbf{Evaluation metric.} We use the normalized final Earth Mover's Distance (EMD) as our performance metric in simulation, defined as: 
\begin{align}
    \frac{D_{EMD}(P^d_{0}, P^d_{g}) - D_{EMD}(P^d_{T}, P^d_{g})} { D_{EMD}(P^d_{0}, P^d_{g})}
    \label{eq-metric}
\end{align}
where $P^d_{0}$, $P^d_{T}$, $P^d_{g}$ are the ground-truth dough point clouds at initialization, final timestep, and the target, respectively. 

\textbf{Baselines.} We consider several baselines in simulation. First, we compare our policy with a  model-free RL baseline trained with partial point clouds as input: Soft Actor Critic (SAC) \cite{haarnoja2017soft}. Both the actor and the critic in SAC use the same inputs and the same encoder as our method. We train the SAC agent for $1$ million timesteps and average the performance over $4$ random seeds.
Second, we compare our trajectory optimizer with the Cross Entropy Method (CEM). Both optimizers (our optimizer and the CEM baseline) operate on the ground-truth state. Details on the hyperparameters used in our experiments are shown in Appendix~\ref{app-hyper}.

\subsection{
Simulation experiments
}
We evaluate all methods on 10 held-out configurations.
The results are shown in \fig{fig-main-result1}. First, we see that our trajectory optimizer (``Diff-Reset (Ours)") outperforms the CEM baseline by a wide margin, highlighting the advantage of our reset module and the gradient information. We also see that our policy (``Diff-Reset-BC (Ours)"), trained using behavior cloning on the demonstration data (generated from trajectory optimization),  outperforms the SAC agent trained using model-free RL.

To further demonstrate the generalization power of our policy, we show the performance of our policy on a larger set of held-out configurations in \fig{fig-heatmap}. The convex hull of the training data is shown in dotted grey lines, and the evaluation configurations are circles whose colors correspond to the normalized performance. As the result suggests, our policy can extrapolate to unseen configurations, despite a few failure cases where the target size is too small compared to the dough size.
\subsection{Ablation studies}
We first investigate the effects of rolling timesteps and the number of reset modules used in our task. For a fixed horizon $T=100$, we vary the number of reset modules from $0$ to $3$ and divide the trajectory into equal numbers of timesteps for rolling, i.e. $t^{reset}_i = \lfloor \frac{100}{N_{reset}}\rfloor \cdot i$.
\tbl{tab-reset} shows the normalized performance of using different number of resets in a trajectory. Most trajectories with resets outperform the one without any reset, highlighting the importance of multiple initializations, and two-stage rolling with one reset in between performs the best for our task.
\label{exp-ablation}
\begin{table}[ht]
\vspace{-2mm}
\centering
\caption{\small Effect of the number of resets in a fixed-horizon trajectory evaluated on 10 held-out configurations.
}
\label{tab-reset}
\begin{tabular}{ |c|c|c|c| }
\hline
No-Reset & 1-Reset & 2-Reset & 3-Reset \\
 \hline
 $0.48 \pm 0.15$ & $\boldsymbol{0.70 \pm 0.05}$ & $0.57 \pm 0.05$ & $0.42 \pm 0.01$\\
 \hline
\end{tabular}
\vspace{-2mm}
\end{table}
We Also consider the following ablations to our novel trajectory optimizer: 
\begin{itemize}
    \item No-Reset: optimize the whole trajectory without reset.
    \item Sep-Reset: optimize multiple rolling trajectories separately with a (non-differentiable) reset of tools in b/w.
    \item Learn-Reset: instead of using the reset module, try to learn the actions that moves the tool back via a reset loss $L^{reset}_t = D(X^{tool}_t, X^{reset}_{i})$.
\end{itemize} 
We compare the performance of our method and the ablations over the held-out trajectories in \fig{fig-main-result1} and include the loss curves of an example trajectory in \fig{fig-main-result2}. As the result suggests, No-Reset (optimization without reset) gets stuck in local optima due to the complexity of our task. Sep-Reset (optimizing the rolling trajectories separately) results in the dough being broken in half, showing the importance of propagating gradient information from one trajectory to another. Learn-Reset tries to optimize for two loss functions, one for rolling and the other for resetting the roller. Since the two losses have conflicting gradient directions, one usually dominates the other, and the roller either ignores the reset action or doesn't performing rolling at all. Please see our website for visualizations of the baselines.

\begin{figure}[t]
    \centering
    \includegraphics[width=0.7\linewidth]{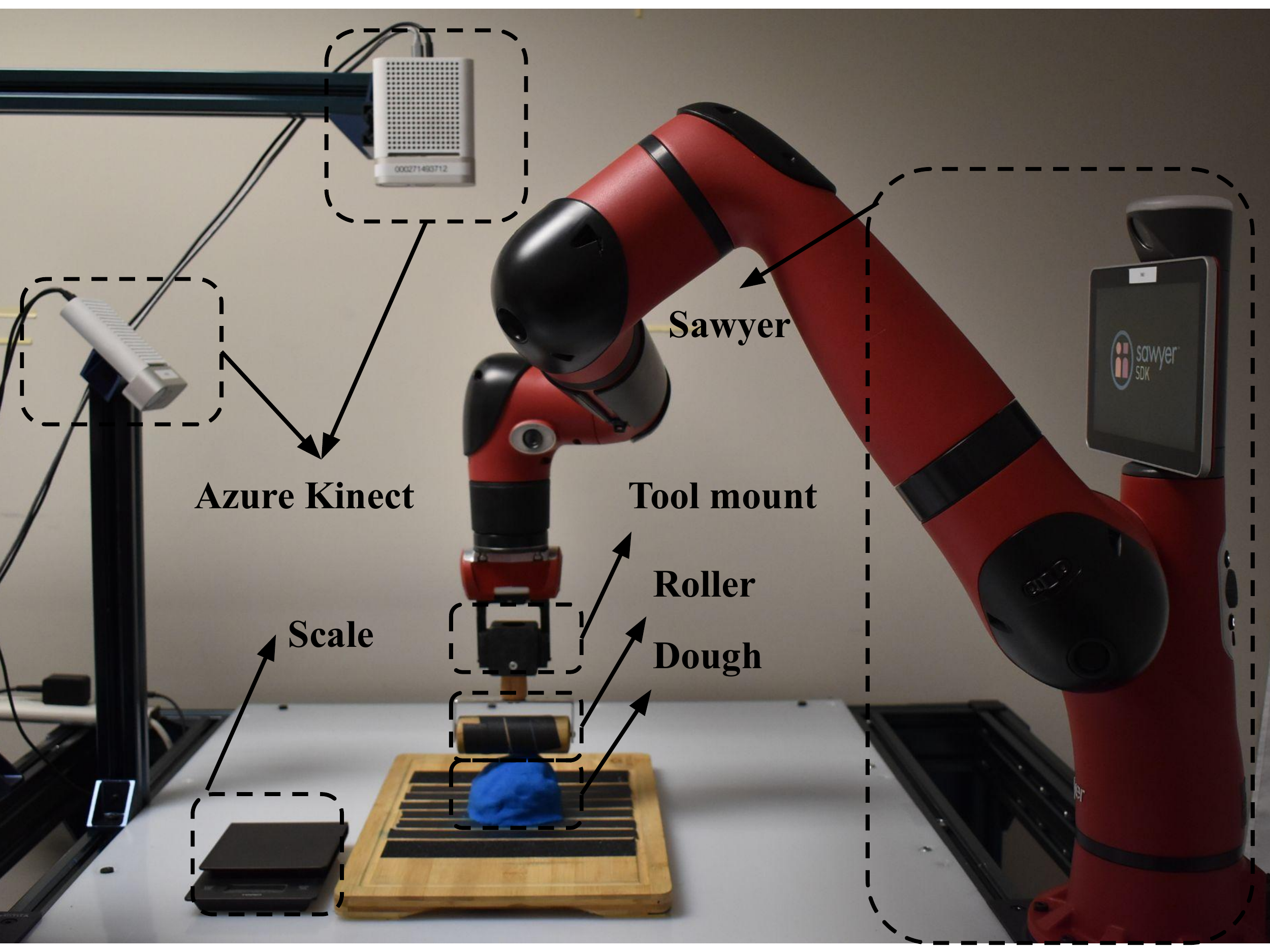}
    \caption{\small Real world workspace setup.
    }
    \label{fig-workspace}
    \vspace{-5mm}
\end{figure}

\begin{figure*}[t]
\vspace{5pt}
     \begin{subfigure}[b]{0.75\textwidth}
         \centering
         \includegraphics[width=0.14\textwidth]{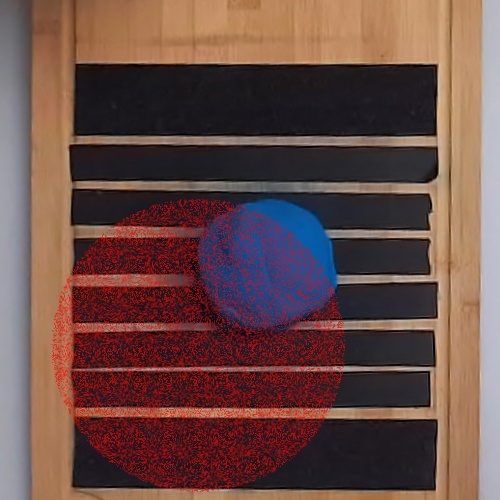}
         \includegraphics[width=0.14\textwidth]{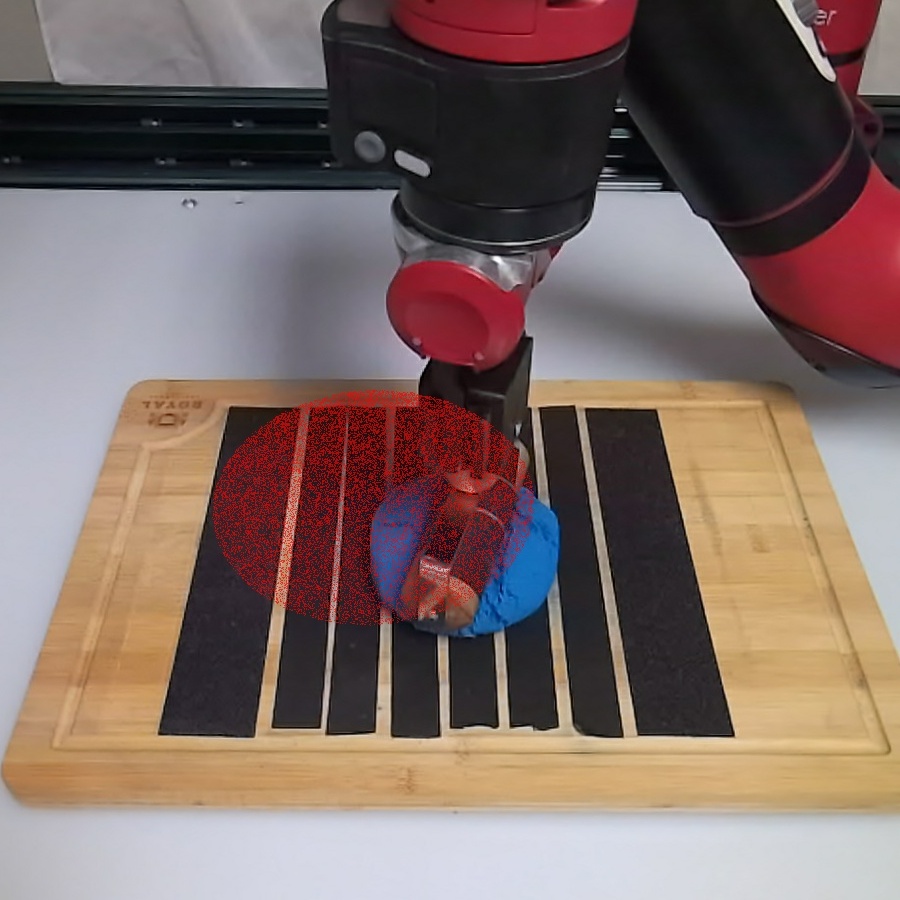}
         \includegraphics[width=0.14\textwidth]{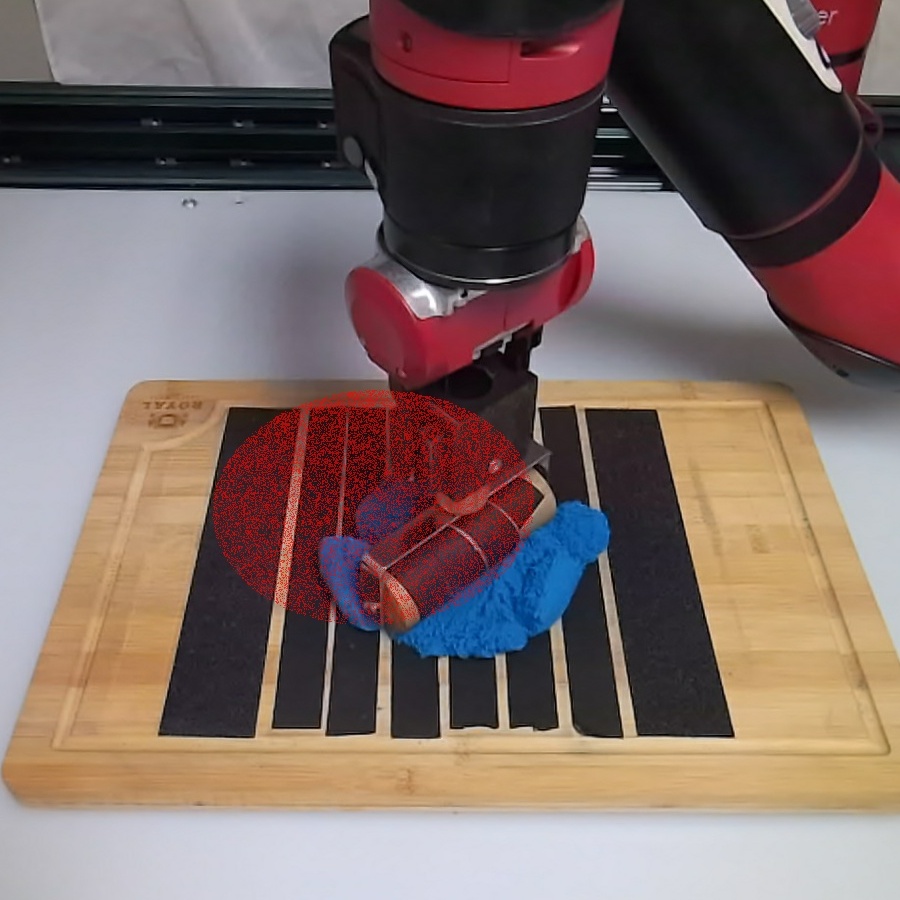}
         \includegraphics[width=0.14\textwidth]{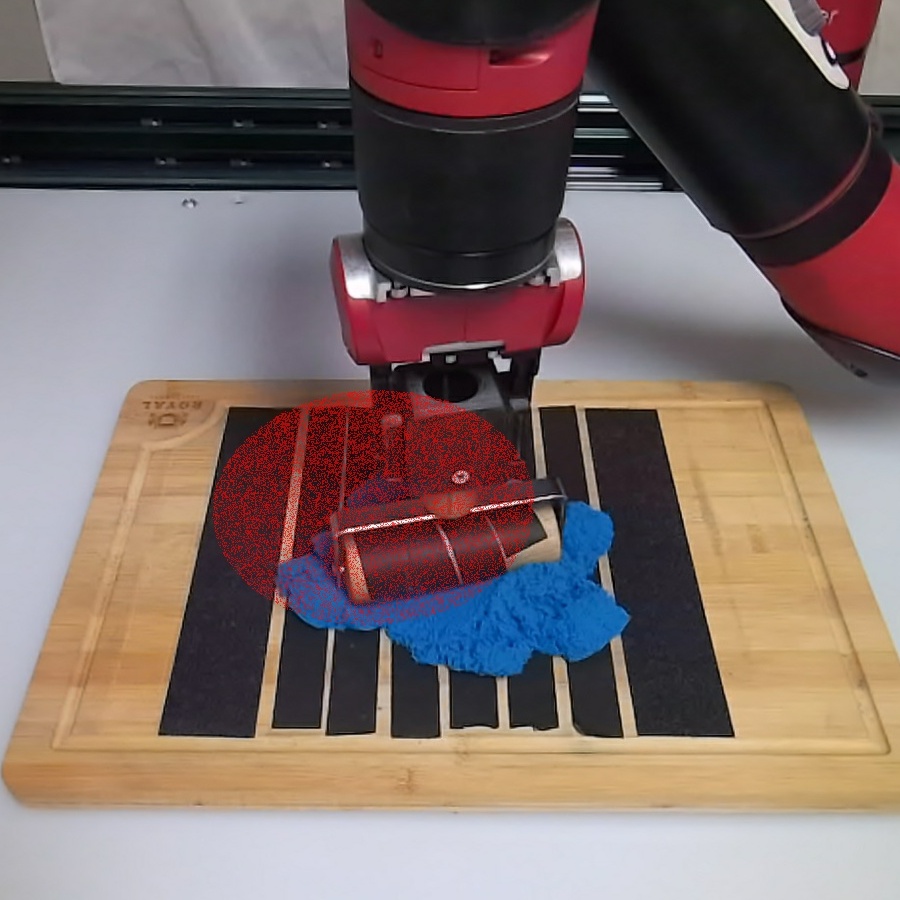}
         \includegraphics[width=0.14\textwidth]{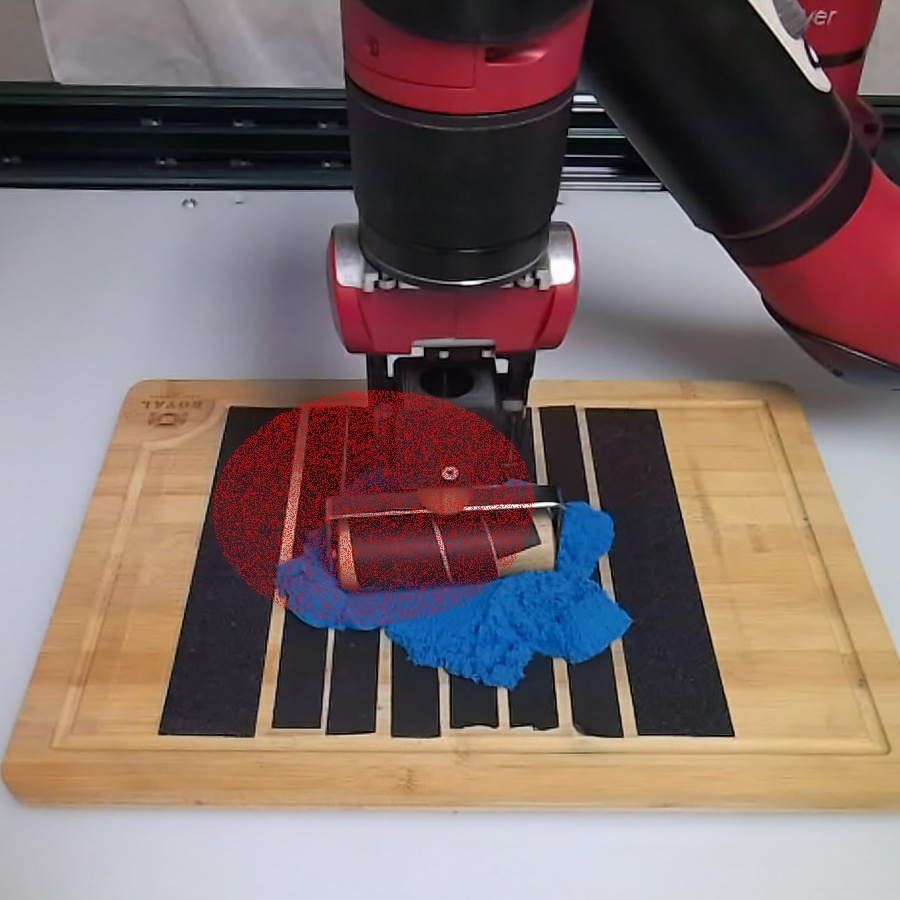}
         \includegraphics[width=0.14\textwidth]{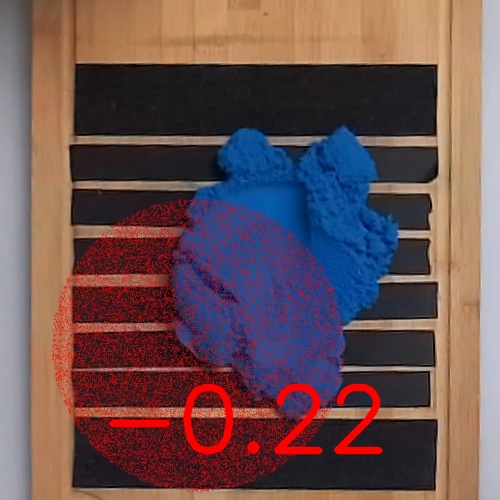}
     \end{subfigure}
     \centering
     \begin{subfigure}[b]{0.75\textwidth}
         \centering
         \includegraphics[width=0.14\textwidth]{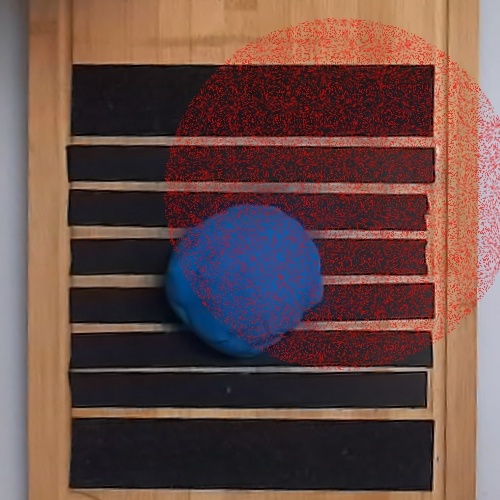}
         \includegraphics[width=0.14\textwidth]{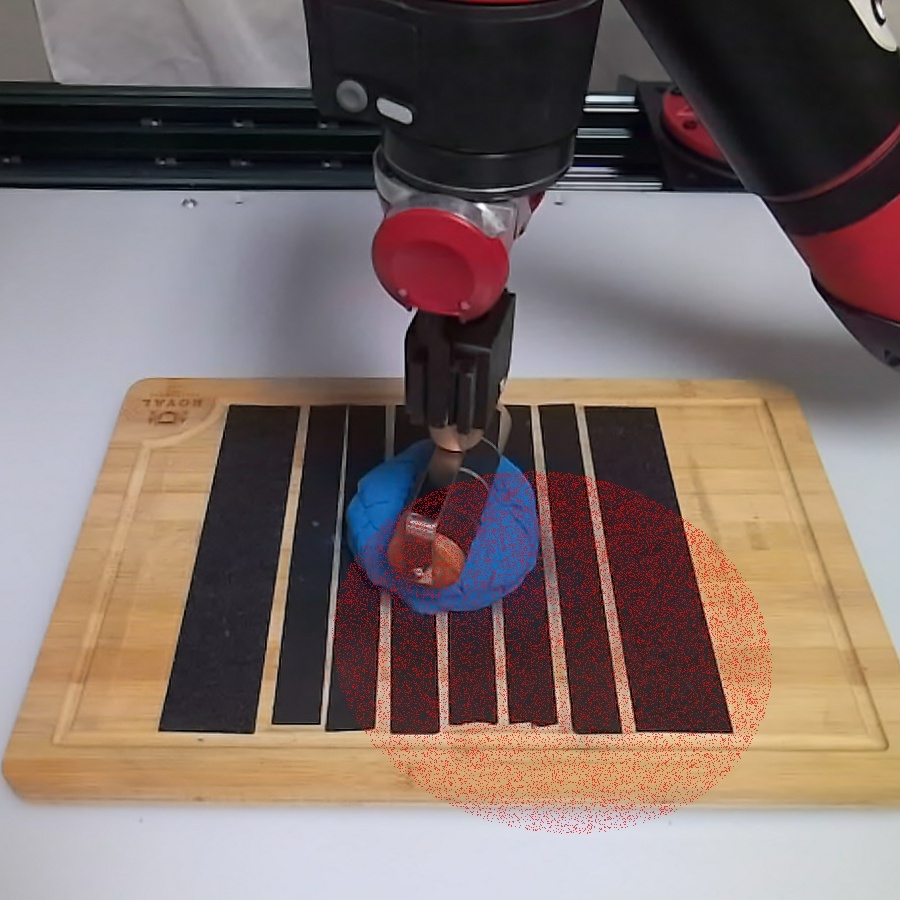}
         \includegraphics[width=0.14\textwidth]{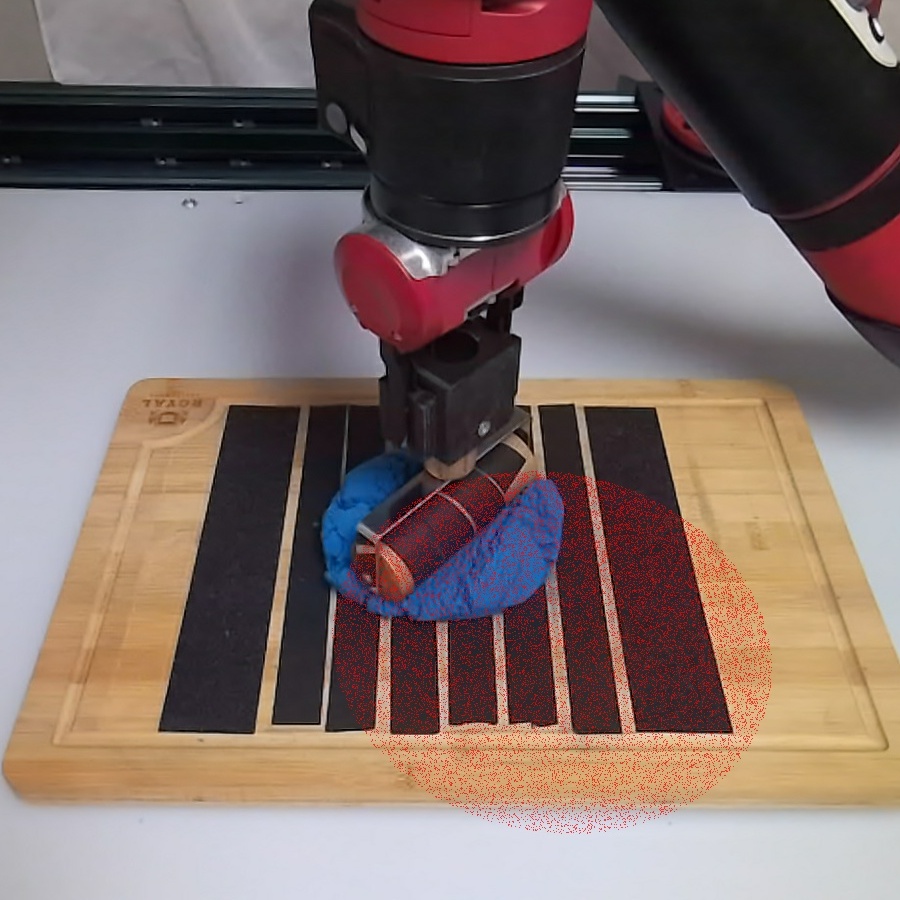}
         \includegraphics[width=0.14\textwidth]{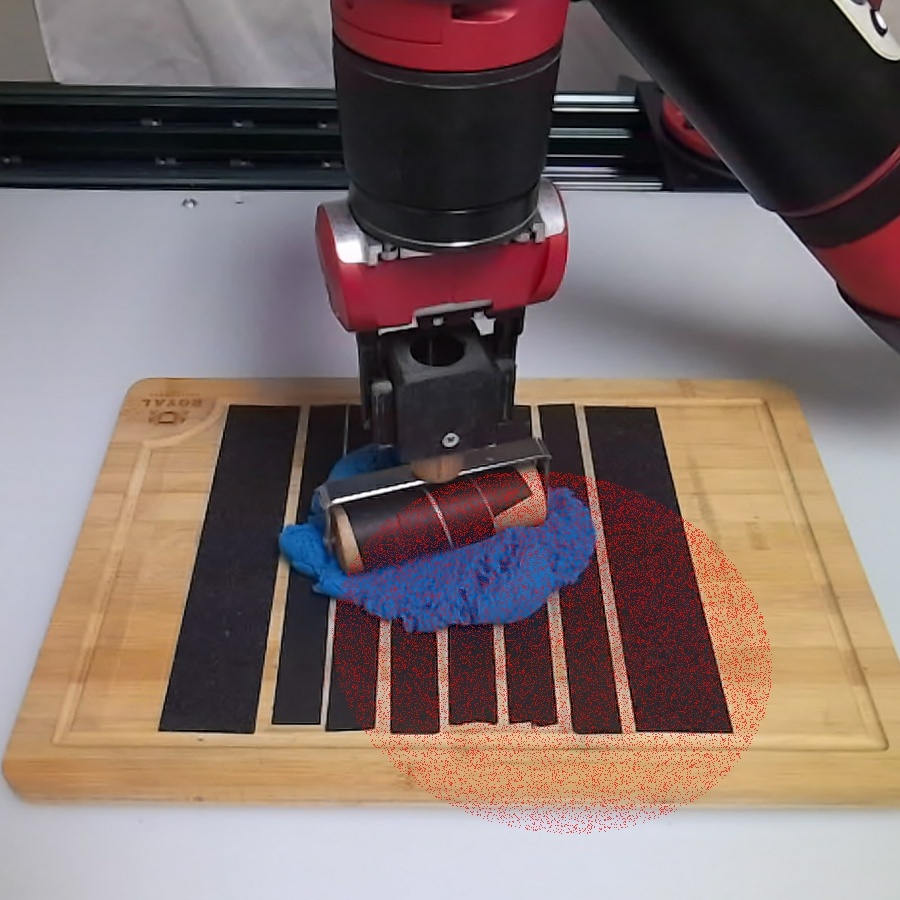}
         \includegraphics[width=0.14\textwidth]{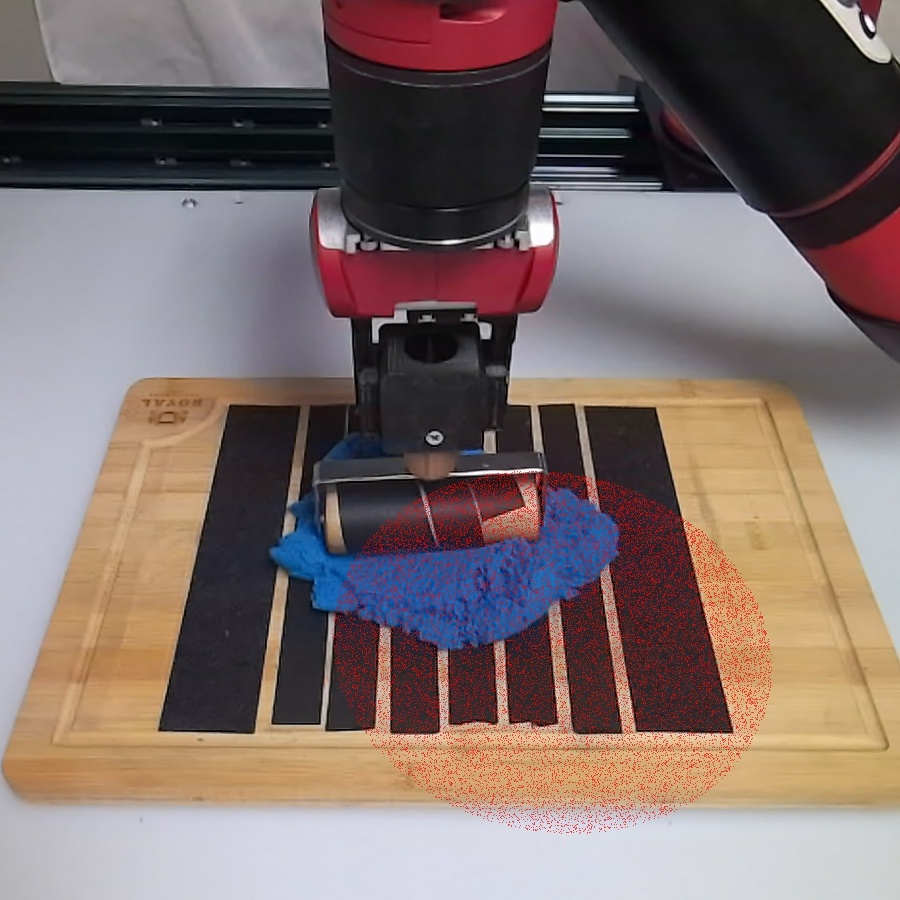}
         \includegraphics[width=0.14\textwidth]{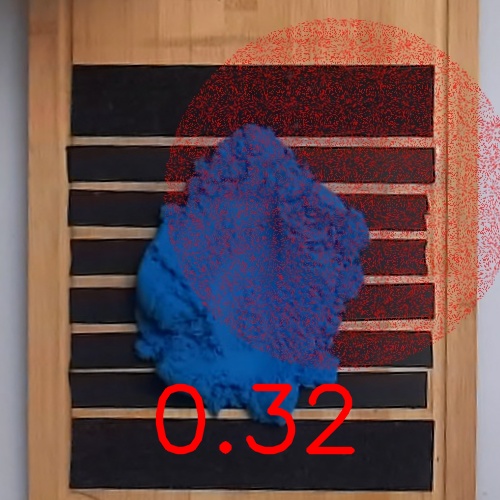}
         \caption{\textbf{Above: Performance of the baseline policy SAC}}
         \label{fig-rollout-sac}
     \end{subfigure}
     \centering
     \begin{subfigure}[b]{0.75\textwidth}
         \centering
         \includegraphics[width=0.14\textwidth]{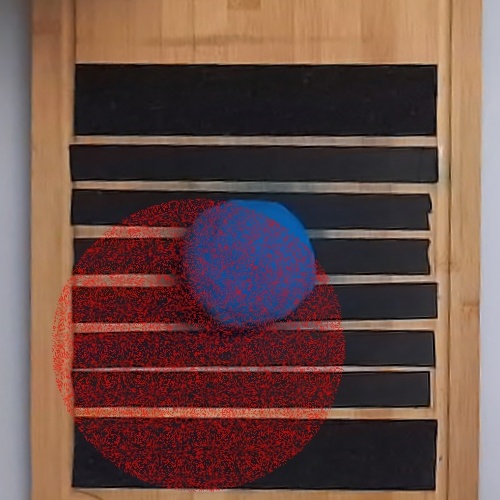}
         \includegraphics[width=0.14\textwidth]{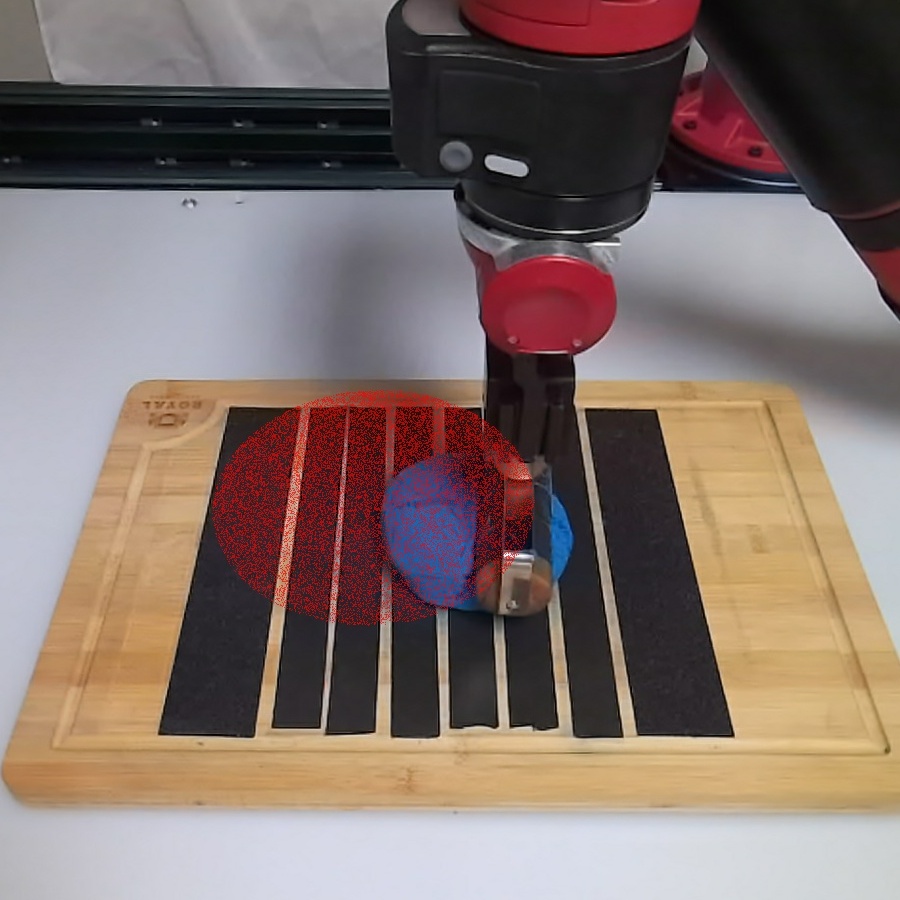}
         \includegraphics[width=0.14\textwidth]{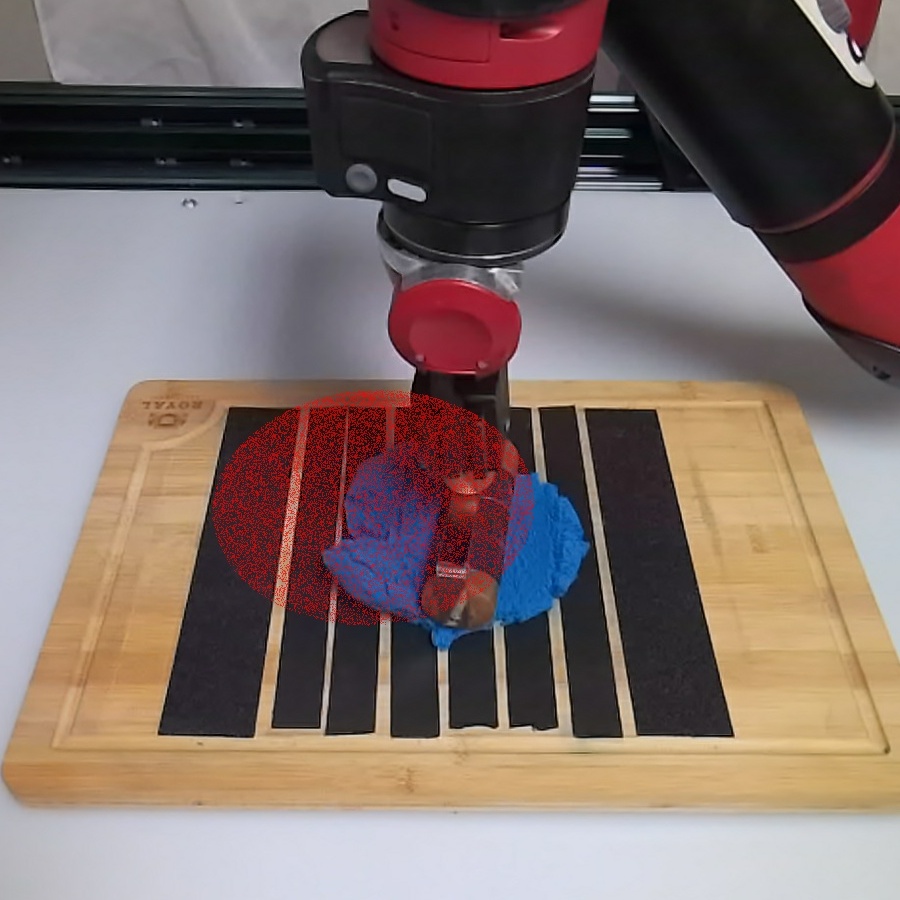}
         \includegraphics[width=0.14\textwidth]{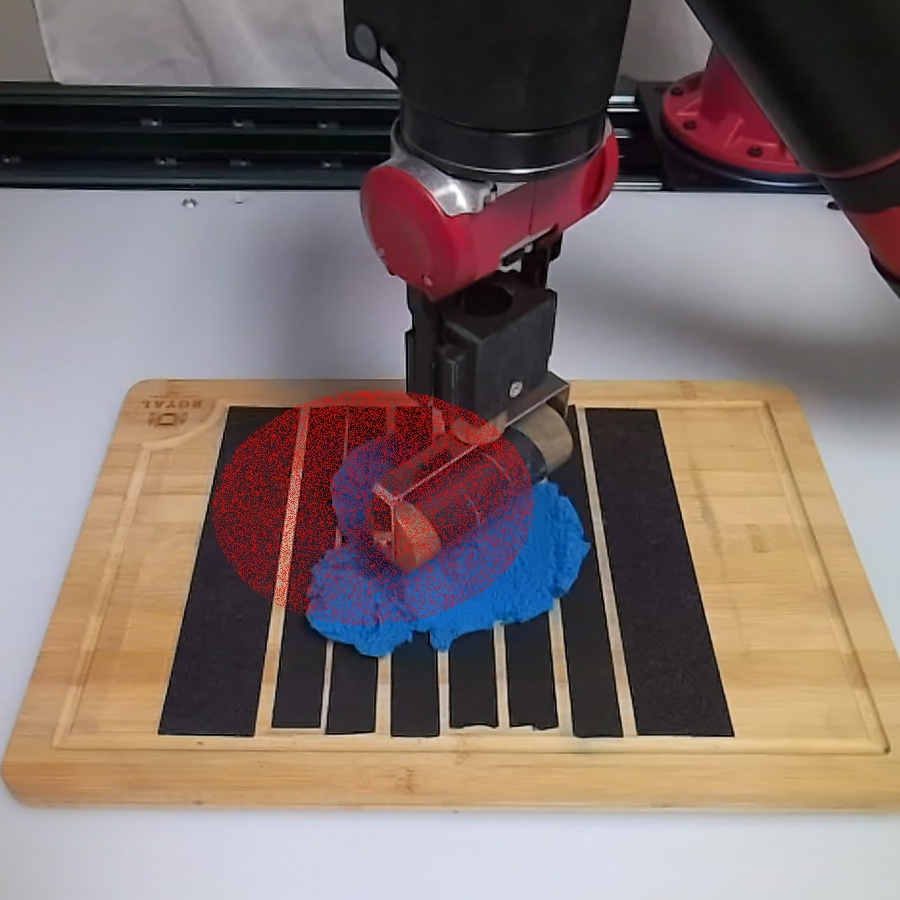}
         \includegraphics[width=0.14\textwidth]{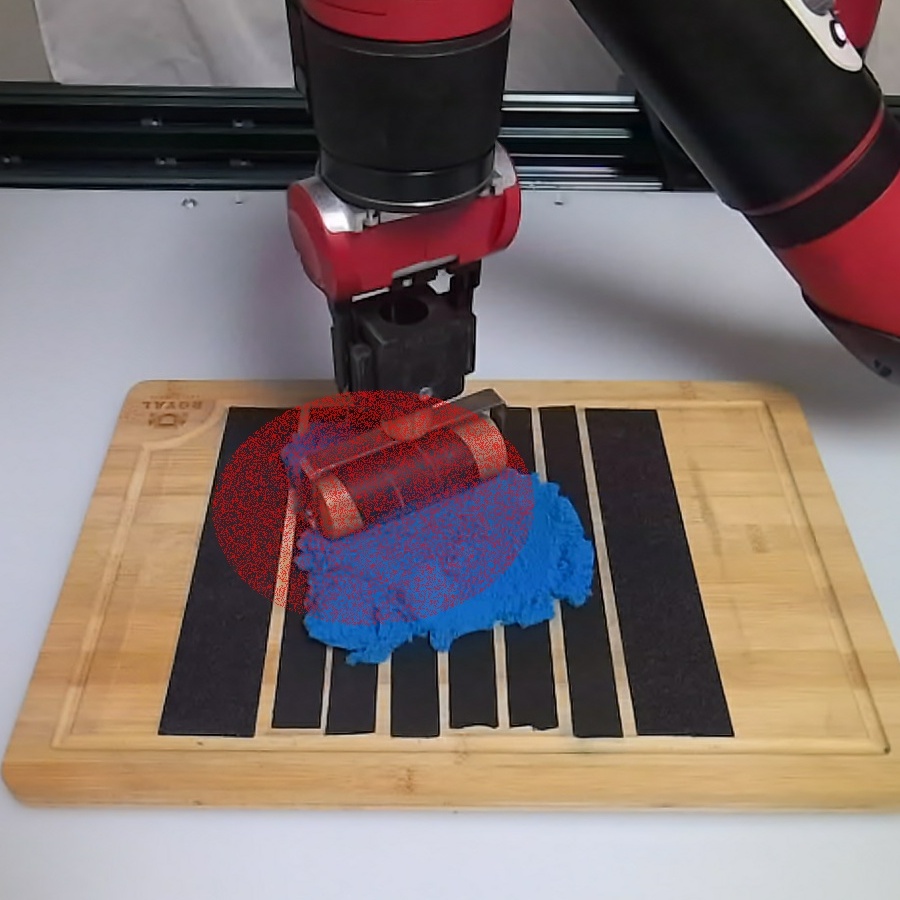}
         \includegraphics[width=0.14\textwidth]{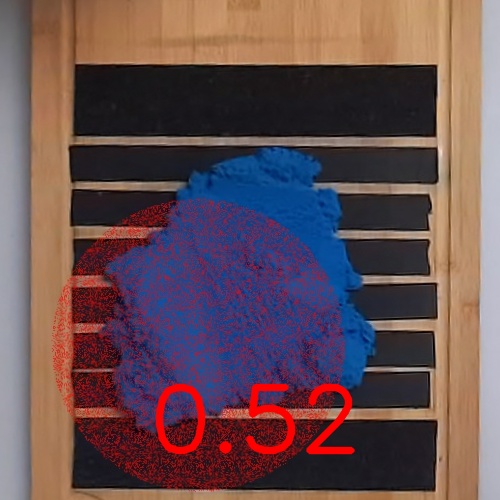}
     \end{subfigure}
     \centering
     \begin{subfigure}[b]{0.75\textwidth}
         \centering
         \includegraphics[width=0.14\textwidth]{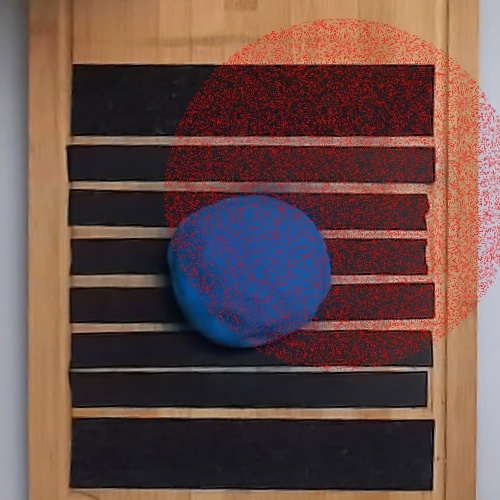}
         \includegraphics[width=0.14\textwidth]{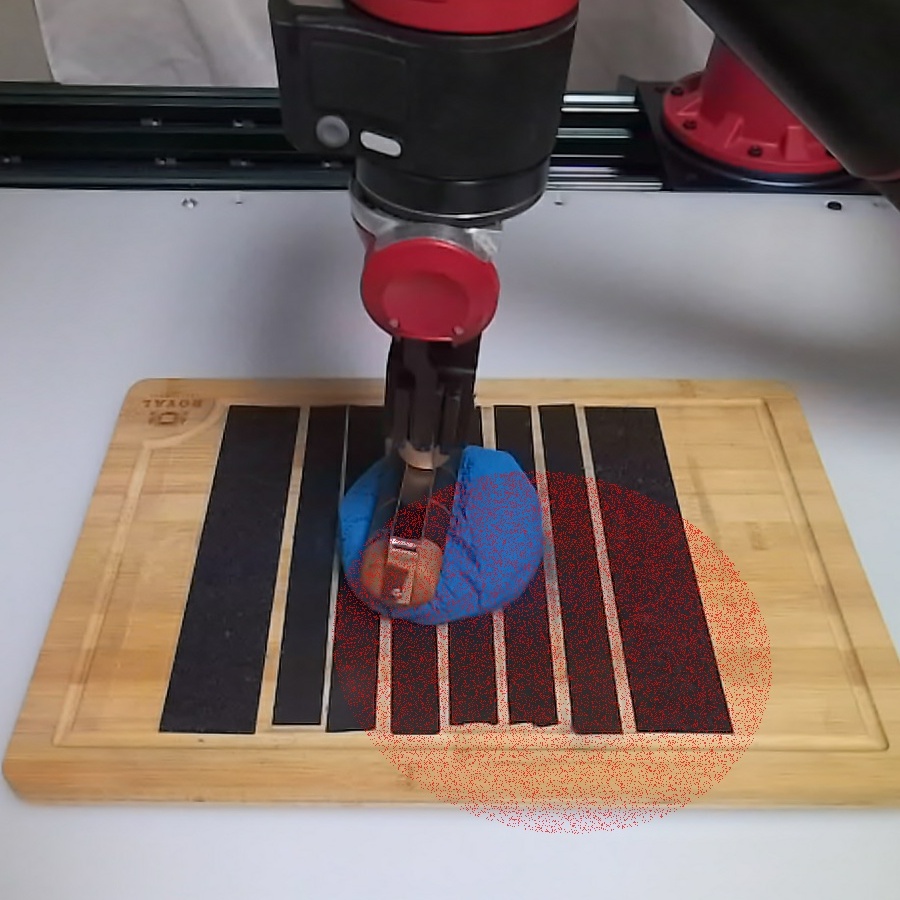}
         \includegraphics[width=0.14\textwidth]{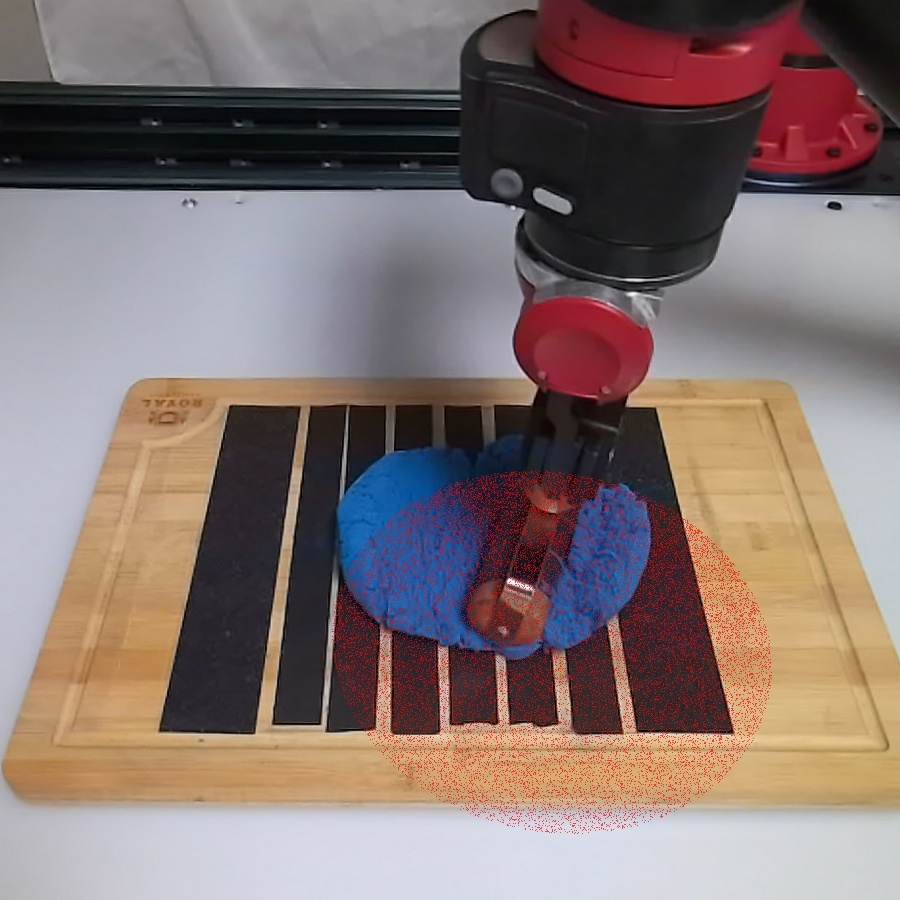}
         \includegraphics[width=0.14\textwidth]{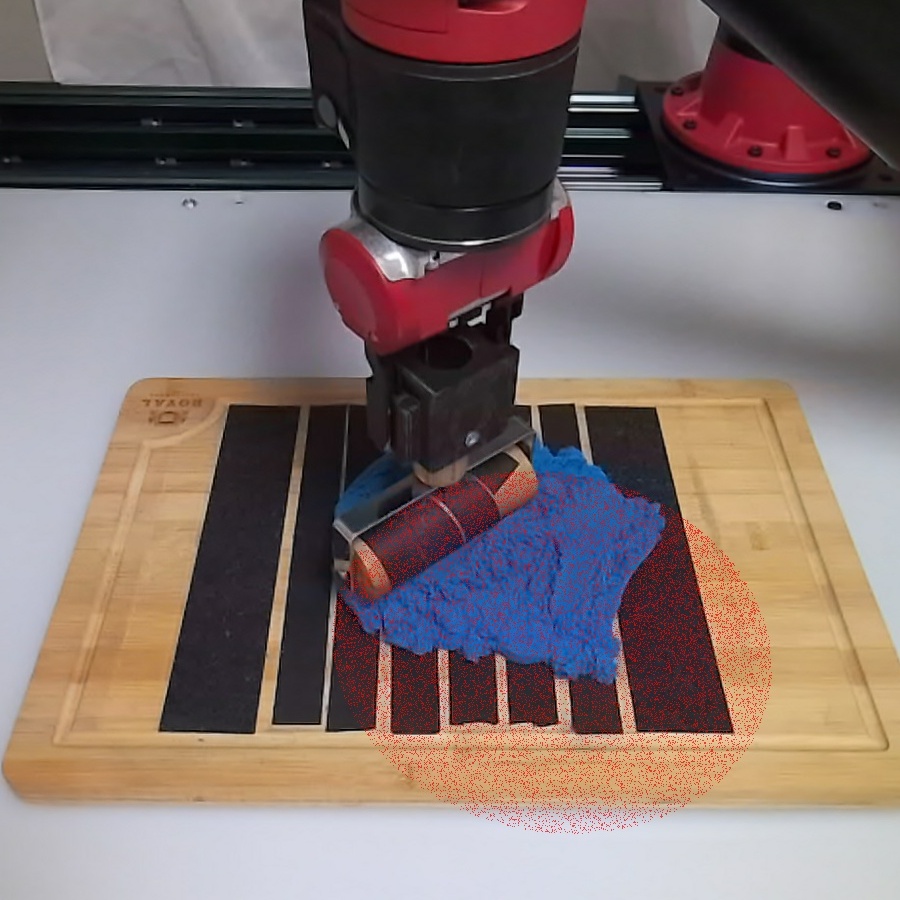}
         \includegraphics[width=0.14\textwidth]{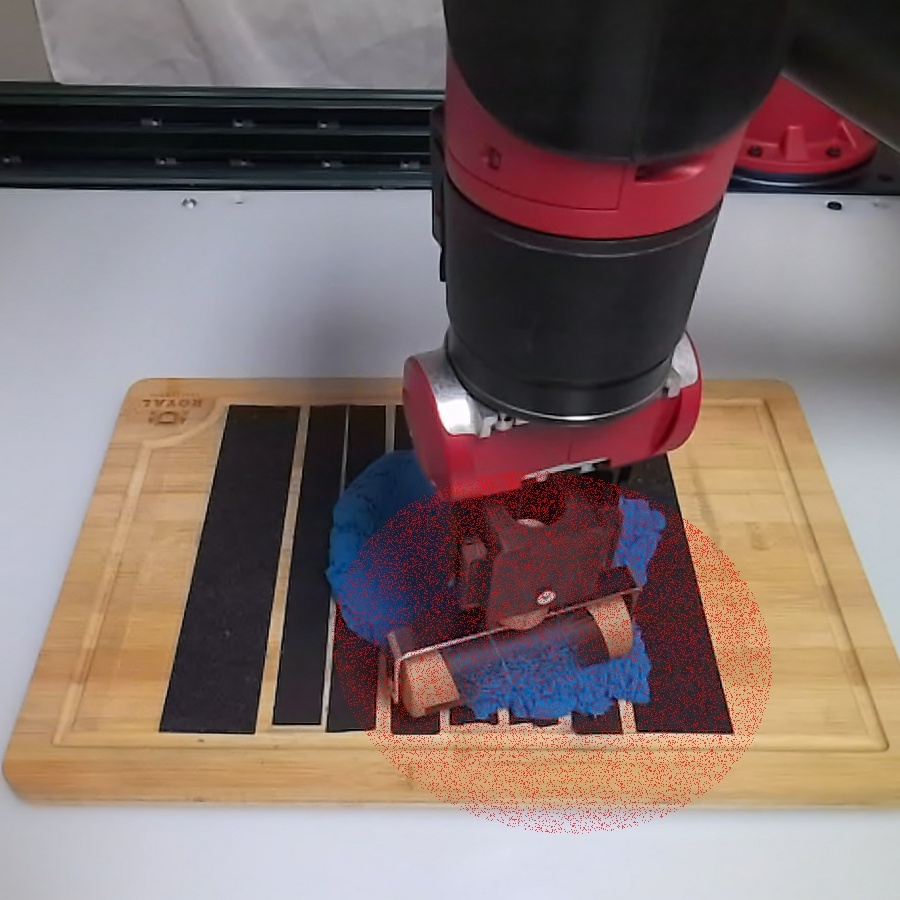}
         \includegraphics[width=0.14\textwidth]{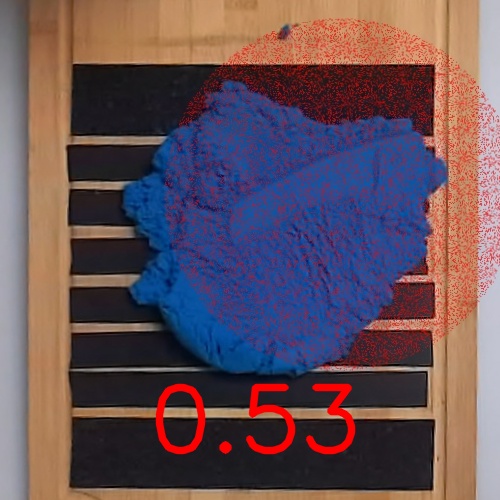}
         \caption{{\textbf{Above: Performance of Diff-Reset-BC (Ours)}}}
         \label{fig-rollout-bc}
     \end{subfigure}
     \caption{\small Example rollouts of SAC~(\ref{fig-rollout-sac}) and our policy~(\ref{fig-rollout-bc}) with goal distance in $6\sim9$ cm. The top-down view of the initial observation and final observation is added to the start and end of the rollouts. The goal shape and normalized performance are overlayed in red.}
     \label{fig-rollout}
     \vspace{-5mm}
\end{figure*}

\subsection{Real world experiments}
\label{exp-realworld}
Our real world setup is shown in \fig{fig-workspace} which includes a Sawyer robot, two Azure Kinect cameras, a tool mount, a roller, and a cutting board. We use Kinetic Sand\footnote{\url{https://kineticsand.com/}} as a proxy for ``dough.'' 
The set of target dough configurations consists of 16 different locations, divided into 3 categories based on the distance between the goal location and the initial location. Since computing the volume of dough in the real world is difficult, we define each of the goal shape to be a flat disk with height zero on the top surface of the cutting board. This results in slight decrease in performance for all the methods because the thickness of the goal shape is not considered. The initial and target configurations of the dough are set in the following way:  we reset the dough by manually rubbing it into a spherical shape and place it in the center of the cutting board. We then sample a goal from the predefined goal set. Last, we input to our policy observations from the side-view camera and execute the current policy until the episode completes. Using the above procedure, our closed-loop policy is able to achieve a control frequency of 2Hz and finish a trajectory in under 1 minute. To evaluate our method, we compute the normalized Chamfer Distance (CD) between the final point cloud and the goal (captured by the top-down camera), similar to Equation~\ref{eq-metric}. We use Chamfer Distance (CD) rather than EMD for the real-world evaluation because EMD requires an equal number of points between the two point clouds, which does not occur for point clouds in the real world. 

We consider three baselines and one ablation in our real world experiments. First, we compare to the SAC agent with the best-performing random seed. Second, we compare to a ``Heuristic": we do a grid search over the rolling depth and rolling length from the initial point cloud observation and execute the same rolling primitive used in prior works \cite{kim2022planning, tokumoto2002deformation}. For a fair comparison, we repeat the rolling primitive twice with the tool reset used by our policy. Third, we compare with a human that observes the overlayed target in real time and is given unlimited time. We also compare to an ablation (``Open-loop") which is an open-loop policy trained on the same demonstration data as our method, but it takes in the initial observation $o_0$ and outputs the entire action sequence $\{\hat a_0, ..., \hat a_{T-1}\}$ in the episode as opposed the action in the next timestep.
\tbl{tab-1} shows the performance of all methods, grouped based on the distance between the initial and final dough configurations. Our policy Diff-Reset-BC outperforms the non-human baselines and ablation in all configurations, and it has the largest improvement over the baselines when the goal is far from the initial location of the dough. Our human baseline is the most dexterous and discovers novel tool uses such as pushing the overextended dough back into the goal region. This highlights the fact that dough manipulation requires a complex action space.
Example rollouts for the SAC agent and Diff-Reset-BC are shown in \fig{fig-rollout}.
Last, we demonstrate the robustness of our policy by varying the size of the dough. We consider 3 different sizes quantified by the weight of the dough, and we scale the radius of the target shape based on the initial dough size. \tbl{tab-2} shows the quantitative results. Diff-Reset-BC outperforms the non-human baselines for all dough sizes and retains a high performance across different sizes. Although the SAC baseline performs on par with our method for small dough, its performance degrades quickly as the dough size increases.
\begin{table}[t]
\centering
\begin{tabular}{ |p{1.5cm}|c|c|c| }
\hline
 & $0\sim3$ cm & $3\sim6$ cm & $6\sim9$ cm \\
 \hline
 Human & $0.72 \pm 0.02$  & $0.63 \pm 0.10$ & $0.67 \pm 0.02$\\
 \Xhline{3\arrayrulewidth}
 SAC & $0.52 \pm 0.04$  & $0.41 \pm 0.06$ & $0.11 \pm 0.20$\\
 \hline
 Heuristic & $0.43 \pm 0.03$ & $0.27 \pm 0.25$ & $0.19 \pm 0.10$ \\ 
 \hline
 Open-loop & $0.59 \pm 0.04$ & $0.26 \pm 0.41$ & $-0.08 \pm 0.45$\\
 \hline
 Diff-Reset-BC (ours) & $\boldsymbol{0.62 \pm 0.07}$ & $\boldsymbol{0.45 \pm 0.06}$ & $\boldsymbol{0.51 \pm 0.04}$\\
 \hline
\end{tabular}
\caption{\small Rolling performance over different target distances. Each entry is evaluated on 4 sampled targets.}
\label{tab-1}
\vspace{-5mm}
\end{table}
Finally, we perform paired samples t-tests~\cite{student1908probable} to statistically compare the performances of our closed-loop policy and each baseline. After applying Bonferroni correction~\cite{rupert2012simultaneous} with $\alpha=0.013$, we find significant differences in performance between Diff-Reset-BC (M=0.51, SD=0.11) and:
\vspace{-1mm}
\begin{itemize}
    \item SAC (M=0.30, SD=0.28); $t(19)=3.6$, $p=1\mathrm{e}{-3}$
    \item Heuristic (M=0.30, SD=0.21); $t(19)=5.6$, $p=2\mathrm{e}{-5}$
    \item Human (M=0.66, SD=0.08); $t(19)=-4.3$, $p=3\mathrm{e}{-4}$
\end{itemize} 
\vspace{-1mm}
as well as marginally significant differences between ours and Open-loop (M=0.28,SD=0.39); $t(19)=2.7$, $p=0.013$.

\begin{table}[t]
\centering
\begin{tabular}{ |p{1.5cm}|c|c|c| }
\hline
  & Small(240g) & Medium(360g) & Large(480g)\\ 
 \hline
 Human & $0.64 \pm 0.01$  & $0.72 \pm 0.02$ & $0.71 \pm 0.01$\\
 \Xhline{3\arrayrulewidth}
 SAC & $\boldsymbol{0.44 \pm 0.12}$  & $0.34 \pm 0.21$ & $0.0 \pm 0.36$\\
 \hline
 Heuristic & $0.33 \pm 0.10$ & $0.31 \pm 0.13$ & $0.21 \pm 0.38$ \\ 
 \hline
 Open-loop & $0.35 \pm 0.35$ & $0.26 \pm 0.41$ & $0.29 \pm 0.36$\\
 \hline
 Diff-Reset-BC (ours) & $\boldsymbol{0.44 \pm 0.09}$ & $\boldsymbol{0.53 \pm 0.09}$ & $\boldsymbol{0.51 \pm 0.17}$\\
 \hline
\end{tabular}
\caption{\small Rolling performance over different dough sizes. Each entry is evaluated on 4 sampled targets.}
\label{tab-2}
\vspace{-5mm}
\end{table}
\section{CONCLUSION}
We introduce a system for closed-loop dough manipulation from high dimensional inputs. Our novel gradient-based trajectory optimizer leverages a differentiable reset module and can optimize an entire multistage trajectory to avoid local optima. We use the trajectory optimizer to generate high-quality demonstration data in a differentiable simulator, which we later use to train a policy via Behavioral Cloning. Our policy trained on partial point cloud is directly transferred to the real world without any fine-tuning. In both simulation and real world experiments, we demonstrate that our method outperforms other approaches and generalizes to different dough sizes and configurations. 
\printbibliography
\newpage
\appendices




\section{Experiment Details}
\label{app-2}

\subsection{Network Architectures}
\label{app-policy}
Our policy consists of a standard PointNet++ \cite{qi2017pointnetplusplus} encoder and a three-layer MLP with hidden dimensions $[1024, 512, 256]$ and ReLU activations. Before the points are inputted to the encoder, we use a one-hot encoding to differentiate points that belong to the tool v.s. points that belong to the current dough observation v.s. points that belong to the target dough shape. We use PyTorch Geometric's \cite{Fey/Lenssen/2019} implementation of PointNet++ and use the following modules in our encoder.

\begin{verbatim}
SAModule(0.5,0.05,MLP([3+3,64,64,128]))
SAModule(0.25,0.1,MLP([128+3,128,128,256]))
GlobalSAModule(MLP([256+3,256,512,1024]))
\end{verbatim}

\subsection{Hyperparameters}
\label{app-hyper}
We perform a grid search over the different hyperparameters used in our SAC \cite{haarnoja2017soft} and CEM baselines. The results are denoted in \tbl{app-hyper}. The numbers in a list denotes the values that we search over, and the bolded numbers are the ones used in generating the final results.

\begin{table}[ht]
    \centering
    \caption{\small Hyperparameters used in SAC and CEM.}
    \begin{tabular}[width=0.4\linewidth]{|l|c|}
    \hline
    Parameters & Values \\
    \hline
    $\alpha$ & 0.2 \\
    \hline
    Tune alpha & [\textbf{True}, False] \\
    \hline
    $\sigma$ & 0.1 \\
    \hline
    lr & [3e{-3}, 3e{-4}, \textbf{3e{-5}}] \\
    \hline
    Batch size & [5, \textbf{10}] \\
    \hline
    $\gamma$ & 0.99 \\
    \hline
    $\tau$ & 0.005 \\
    \hline
    $\lambda_{contact}$ & [1e-3, \textbf{10}, 20] \\ 
    \hline
    Training steps & 1000000 \\
    \hline
    \end{tabular}
    \begin{tabular}[width=0.4\linewidth]{|l|c|}
    \hline
    Parameters & Values \\
    \hline
    Horizon & [1, 5, \textbf{10}, 20] \\
    \hline
    Pop. size & [50, \textbf{100}] \\
    \hline
    Max iter & [\textbf{10}, 20] \\
    \hline
    Elites & 10 \\
    \hline
    \end{tabular}
    \label{tab-hyper}
\end{table}

\end{document}